\documentclass{article}

\usepackage{microtype}
\usepackage{graphicx}

\usepackage{caption}
\usepackage{subcaption}
\usepackage{booktabs} 
\usepackage{siunitx}
\usepackage{hyperref}
\usepackage{svg}

\usepackage{makecell}

\usepackage{dsfont}
\usepackage[accepted]{icml2023}

\usepackage{amsmath}
\usepackage{amssymb}
\usepackage{mathtools}
\usepackage{amsthm}
\usepackage{graphics}
\usepackage{xcolor,pifont}
\usepackage[capitalize,noabbrev]{cleveref}

\usepackage{listings}
\theoremstyle{plain}

\theoremstyle{definition}

\theoremstyle{remark}

\usepackage[textsize=tiny]{todonotes}

\icmltitlerunning{Posterior Sampling for Deep Reinforcement Learning}

\DeclareMathOperator*{\argmax}{argmax} 

\newcommand*\colourcheck[1]{%
  \expandafter\newcommand\csname #1check\endcsname{\textcolor{#1}{\ding{52}}}%
}
\newcommand*\colourmark[1]{%
  \expandafter\newcommand\csname #1mark\endcsname{\textcolor{#1}{\ding{55}}}%
}

\colourcheck{green}
\colourmark{red}
\begin{document}

\twocolumn[
\icmltitle{Posterior Sampling for Deep Reinforcement Learning}



\icmlsetsymbol{equal}{*}

\begin{icmlauthorlist}
\icmlauthor{Remo Sasso}{yyy}
\icmlauthor{Michelangelo Conserva}{yyy}
\icmlauthor{Paulo Rauber}{yyy}
\end{icmlauthorlist}

\icmlaffiliation{yyy}{School of Electronic Engineering and Computer Science, Queen Mary University of London, United Kingdom}

\icmlcorrespondingauthor{Remo Sasso}{r.sasso@qmul.ac.uk}

\icmlkeywords{Reinforcement Learning, Model-Based Reinforcement Learning, Bayesian Reinforcement Learning, Posterior Sampling, Bayesian Neural Networks, Deep Reinforcement Learning}
\vskip 0.3in
]



\printAffiliationsAndNotice{}  

\begin{abstract}
Despite remarkable successes, deep reinforcement learning algorithms remain sample inefficient: they require an enormous amount of trial and error to find good policies.
Model-based algorithms promise sample efficiency by building an environment model that can be used for planning.
Posterior Sampling for Reinforcement Learning is such a model-based algorithm that has attracted significant interest due to its performance in the tabular setting.
This paper introduces Posterior Sampling for Deep Reinforcement Learning (PSDRL), the first truly scalable approximation of Posterior Sampling for Reinforcement Learning that retains its model-based essence.
PSDRL combines efficient uncertainty quantification over latent state space models with a specially tailored continual planning algorithm based on value-function approximation.
Extensive experiments on the Atari benchmark show that PSDRL significantly outperforms previous state-of-the-art attempts at scaling up posterior sampling while being competitive with a state-of-the-art (model-based) reinforcement learning method, both in sample efficiency and computational efficiency.
\end{abstract}

\section{Introduction} 
\label{introduction}

In a typical reinforcement learning problem, an agent interacts with an environment in a sequence of episodes by observing states and rewards and acting according to a policy that maps states to actions. 
A reinforcement learning algorithm seeks a policy that maximizes expected cumulative rewards. Because many problems in healthcare, robotics, logistics, finance, and advertising can be naturally formulated as problems of maximizing a measure of success through a sequence of decisions informed by data, the recent successes of reinforcement learning have attracted significant interest. In particular, the combination of reinforcement learning with artificial neural networks has led to the best computer agents that play games such as Chess and Go \citep{Schrittwieser2019-df}, Dota 2 \citep{berner2019dota}, and StarCraft II \citep{vinyals2019grandmaster}. 

Despite these notable successes, the corresponding reinforcement learning algorithms are remarkably \emph{sample inefficient}: they require an enormous amount of trial-and-error to find good policies. In contrast with games and simulations, real-world applications are heavily constrained by the cost of trial-and-error and available data. As a consequence, sample inefficiency limits the applicability of reinforcement learning. This inefficiency is fundamentally linked with the trade-off between \emph{exploring} an environment in order to learn about potentially better sources of reward and \emph{exploiting} the well-known sources of reward. In the tabular reinforcement learning setting, where the number of states is finite, several (theoretically and often empirically) efficient exploration methods are well understood \citep{osband2013more, agrawal2017optimistic, azar2017minimax,zanette2019tighter,   russo2019worst, menard2021ucb}. However, there are no generally efficient methods that cope with the non-linear function approximation required in non-tabular settings. 

Model-based reinforcement learning methods seek sample efficiency by building an environment model that enables predicting how actions affect the state of the environment and how the states of the environment relate to rewards. Because such a model can be used for planning (searching for a good policy without interacting with the environment), model-based methods have the potential to be substantially more sample efficient than model-free algorithms \citep{Kaiser2019-ts, janner2019trust}, which attempt to find good policies without building a model. Recent work has shown that learning models in latent state space can significantly reduce the computational cost of model-based reinforcement learning \citep{Ha2018-jd, hafner2019learning, hafner2020dream, Schrittwieser2019-df}, allowing its application in environments with high-dimensional state spaces.

Posterior Sampling for Reinforcement Learning is a model-based algorithm that has attracted significant interest due to its strong (theoretical and empirical) performance in the tabular 
setting \citep{osband2013more}. This algorithm represents its knowledge about an environment by a distribution over environment models and repeats the following steps: a single model is drawn from the distribution over models; an optimal policy is found for this model; this policy is used to interact with the environment for one episode; and the resulting data are used to update the distribution over models. Intuitively, exploration decreases as knowledge increases.

This paper introduces Posterior Sampling for Deep Reinforcement Learning (PSDRL), the first truly scalable approximation of Posterior Sampling for Reinforcement Learning that retains its model-based essence. PSDRL encodes a high-dimensional state into 
a lower-dimensional latent state to enable predicting transitions in latent state space for any given action. 
PSDRL represents uncertainty through a Bayesian neural network that maintains a distribution over the parameters of this transition model. This enables sampling a model that can be used for planning, which is accomplished by value function approximation with an artificial neural network. This so-called value network retains information across sampled models, which is crucial for sample efficiency. Completing the algorithm, the data collected by the agent while acting greedily with respect to the value network is regularly used to update the latent state representations and the distribution over the parameters of the model.

Extensive experiments on the Atari benchmark \cite{bellemare2013arcade} show that PSDRL significantly outperforms previous state-of-the-art attempts at scaling up posterior sampling (Bootstrapped DQN
with randomized priors 
\citep{osband2018randomized} and Successor Uncertainties \citep{janz2019successor}). They also show that PSDRL is competitive with a state-of-the-art (model-based) 
reinforcement learning 
method (DreamerV2 \citep{Hafner2020-ry}), both in sample efficiency and computational efficiency.

The remaining text is organized as follows. Section \ref{sec:related_works} relates our contributions to previous works. Section \ref{sec:psdrl} describes PSDRL in technical detail. Section \ref{sec:experiments} presents the experimental protocol and its results. Finally, Section \ref{sec:conclusion} summarizes our contributions and suggests future work.

\section{Related works}
\label{sec:related_works}

\textbf{Model-based reinforcement learning} has historically underperformed in complex environments that require non-linear function approximation when compared with model-free reinforcement learning. However, two model-based methods have recently matched (or surpassed) the performance of model-free methods in the 
common
Atari game playing benchmark (\citet{Schrittwieser2019-df} and \citet{Hafner2020-ry}). This is an important development since planning has the potential to make model-based methods highly sample efficient \citep{Kaiser2019-ts, janner2019trust}.

\textbf{Exploration} methods that cope with the non-linear function approximation required in challenging environments can be based on one of four foundations \cite{badia2020agent57}: domain-specific knowledge, unsupervised policy learning, intrinsic motivation, or posterior sampling. 
\emph{Domain-specific knowledge} methods typically combine human demonstrations, handcrafted features, and heuristics \cite{aytar2018playing, ecoffet2021first}.
Despite their potential to be highly sample efficient in their target environments, adapting these methods to new environments requires significant effort and expertise.
\emph{Unsupervised policy learning} methods encourage agents to acquire a diverse set of \emph{skills} without receiving reward signals \cite{eysenbach2018diversity}. Given such limited feedback, identifying generally useful skills for efficient exploration through reuse and composition is a difficult task.
\emph{Intrinsic motivation} methods aim to encourage (re)visiting states with bonus rewards derived from 
ensembles of Q-functions \citep{chen2017ucb, bai2021principled, tiapkin2022dirichlet},
visitation counts \citep{bellemare2016unifying, rashid2020optimistic}, or
episodic curiosity \citep{savinov2018episodic, badia2019never}.
Although many of these exploration methods are certainly promising and often effective, our focus on posterior sampling is justified by the fact that Posterior Sampling for Reinforcement Learning is the simplest among the potentially scalable and principled methods that is capable of leveraging the strengths of both Bayesian methods and model-based reinforcement learning.


\textbf{Posterior Sampling for Reinforcement Learning} has been scaled up to non-tabular settings by \citet{tziortziotis2013linear} and \citet{fan2021modelbased}. Both works rely on state-action embeddings and a posterior distribution based on Bayesian linear regression. \citet{tziortziotis2013linear} employ a fixed embedding and explore different approximate dynamic programming techniques (such as least-square policy iteration), which makes their algorithm limited to very simple environments. \citet{fan2021modelbased} learn state-action embeddings and use model predictive control \citep{24286} via the cross-entropy method \citep{BOTEV201335} for planning instead of utilizing a value function, also limiting scalability relatively simple environments. 
The main obstacle in scaling up model-based posterior sampling to complex environments is the extreme computational and memory cost of searching for a good policy for a sampled model while retaining too little information from previous searches.

\textbf{Randomized value function} methods are the model-free counterparts of Posterior Sampling for Reinforcement Learning \cite{osband2016deep,osband2018randomized,osband2019deep}. These methods (implicitly) represent their knowledge about the optimal value function by a distribution over value functions and repeat the following steps: a single value function is (implicitly) drawn from the distribution over value functions; a greedy policy is derived from this value function; this policy is used to interact with the environment during one (or more) episodes; and the resulting data are used to (approximately) update the (implicit) distribution over value functions. In this context, \citet{osband2016generalization} approximate randomized least-square value iteration.
\citet{engel2003bayes,engel2005reinforcement} employ Gaussian processes trained via temporal-difference learning to induce a distribution over Q-functions. Similarly, 
\citet{8503252}, \citet{o2018uncertainty}, and \citet{janz2019successor} rely on Bayesian linear regression on top of learned state embeddings, while \citet{flennerhag2020temporal} approximate a distribution over temporal differences. However, explicitly maintaining and accurately updating a distribution that represents knowledge about the optimal value function is generally infeasible \cite{osband2019deep}, and natural approximations of this approach may fail to realize its potential \cite{janz2019successor}. This is partially due to the fact that randomized value function methods
are often based on fitted value iteration, which provides unreliable and noisy learning targets (\citealp{NEURIPS2019_c2073ffa, kumar2020discor}). In contrast with randomized value functions, model-based posterior sampling methods have the potential advantage of separating uncertainty about the environment from uncertainty about the optimal value function (or policy).

\begin{figure*}[h]
    \centering
    \vskip 0.2in
    \hfill
    \begin{subfigure}[t]{0.42\linewidth}
        \centering
        \includegraphics[width=0.77\linewidth]{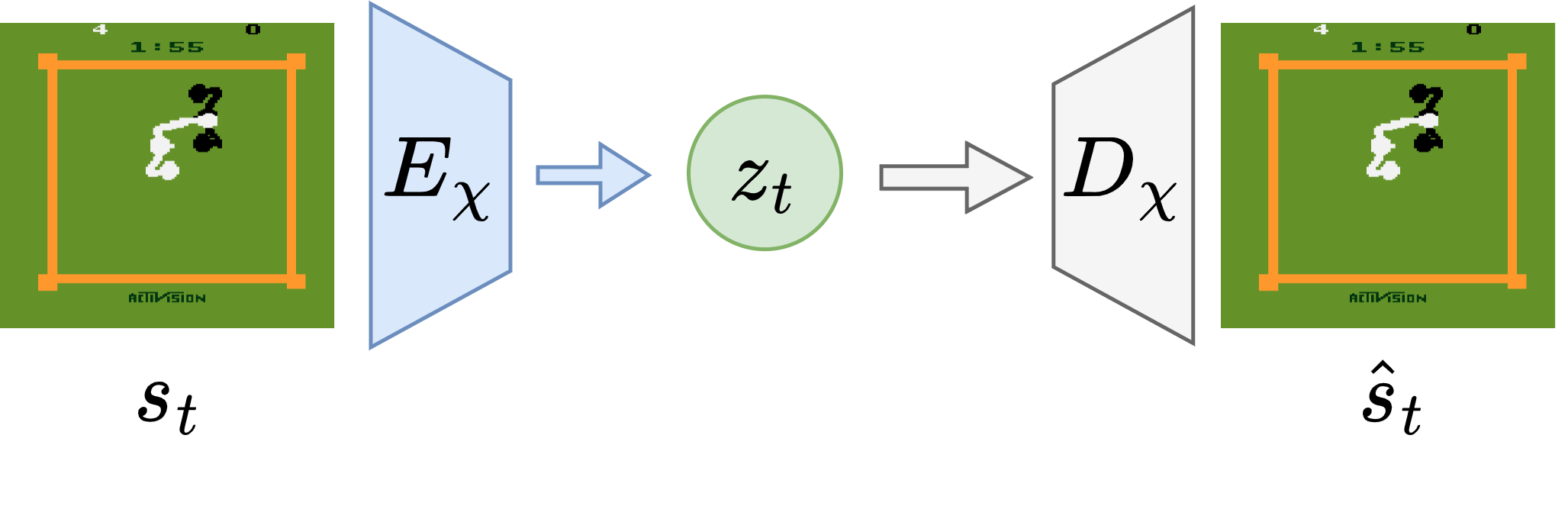}
        \caption{}
    \end{subfigure}
    \hfill
    \begin{subfigure}[t]{0.49\linewidth}
        \centering
        \includegraphics[width=0.77\linewidth]{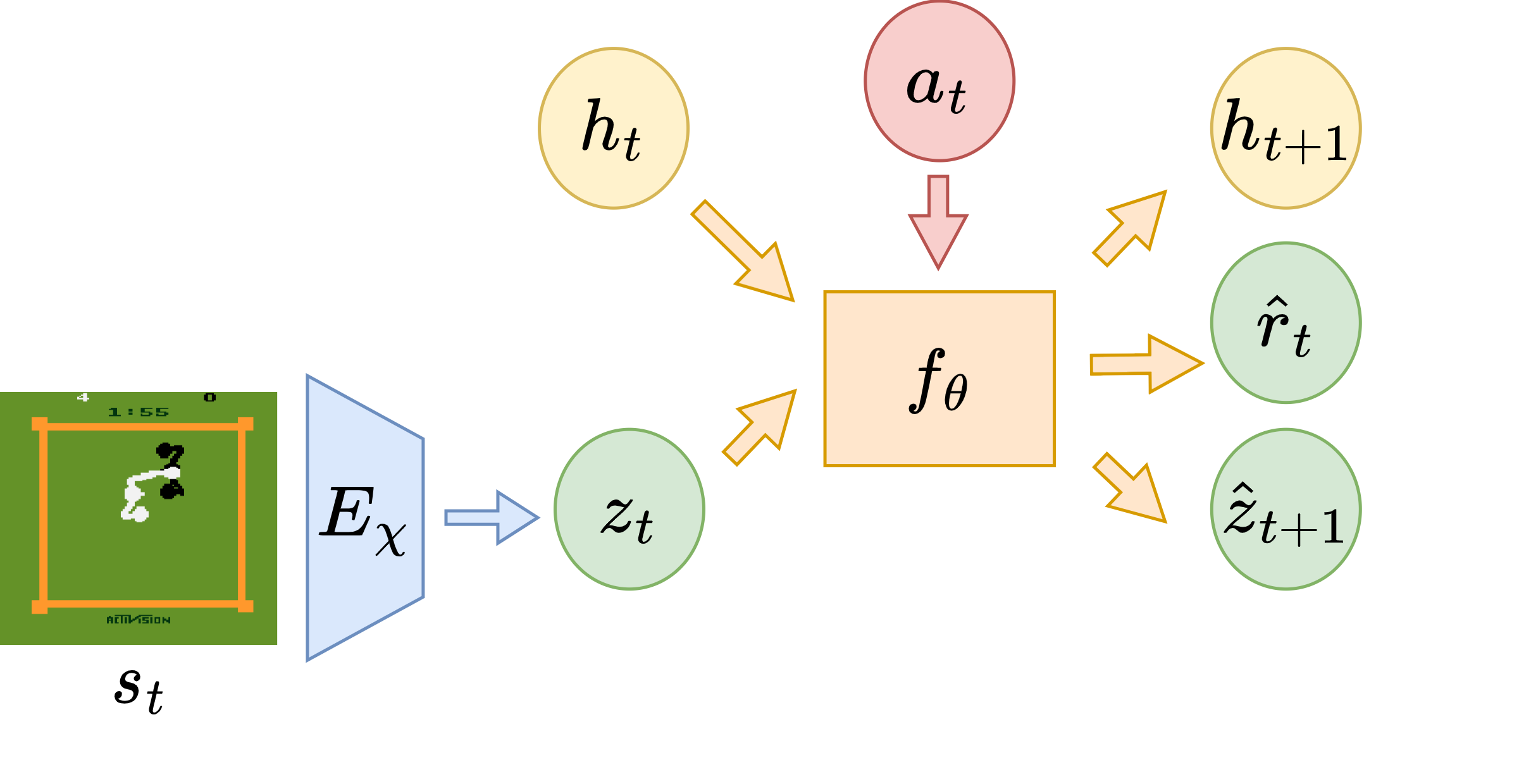}
        \caption{}
    \end{subfigure}
    \hfill
    \caption{(a) The autoencoder learns latent state representations $z_t$ for true environment states $s_t$ through reconstruction. (b) The forward model learns to predict rewards $r_t$ and next latent states $z_{t+1}$ given the current latent states $z_t$, actions $a_t$, and previous hidden states $h_t$.}
    \label{fig:diagram}
\vskip -0.1in
\end{figure*}

\section{Posterior Sampling for Deep Reinforcement Learning}
\label{sec:psdrl}

This section introduces the PSDRL algorithm, which approximates Posterior Sampling for Reinforcement Learning while retaining its model-based essence. Section \ref{sec:background} introduces our notation. Section \ref{sec:dynamics_model} describes the latent state space transition model. Section \ref{sec:posterior_distributions} explains how uncertainty over these models is represented. Section \ref{sec:posterior-sampling} details planning based on sampled models. Finally, Section \ref{sec:justification} justifies important implementation choices and reports potential pitfalls in scaling up model-based posterior sampling.

\subsection{Preliminaries}
\label{sec:background}

We consider a deterministic sequential decision-making problem where, at a given time step $t$, an agent observes a state $s_t$ from a state space $\mathcal S$ and chooses an action $a_t$ from an action space $\mathcal A$ using a policy $\pi : \mathcal S \to \mathcal A$, which results in a next state $s_{t+1} = T(s_t, a_t)$ and reward $r_t = R(s_t, a_t)$. The goal of the agent is to find a  policy $\pi$ that maximizes the return $G(\pi) = \sum_{t=0}^{\infty}{\gamma^t r_t}$, where $\gamma \in [0,1)$ is a discount factor.
The action-value $Q_\pi(s,a)$ is the return from following policy $\pi$ starting from state $s$ and action $a$, while the value $V_\pi(s)$ is given by  $V_\pi(s) = Q_\pi(s,\pi(s))$.

\subsection{Transition model}
\label{sec:dynamics_model}

Inspired by the recent remarkable successes of latent space models in model-based reinforcement learning \citep{Ha2018-jd, Schrittwieser2019-df, hafner2020dream, Hafner2020-ry}, PSDRL employs a latent space transition model (Fig. \ref{fig:diagram}) composed of the following components (subscripts denoting parameters): autoencoder $(E_{\chi}, D_{\chi})$; forward model $f_{\theta}$; and termination model $\omega_{\eta}$ .

The \emph{autoencoder} is convolutional  \citep{masci2011stacked}. For a given state $s_t$, the encoder $E_{\chi}$ produces a latent state $z_t = E_{\chi}(s_t) $ that may be decoded by the decoder $D_{\chi}$ into a state estimate $\hat{s}_t = D_{\chi}(z_t)$. Encoding a state space into a lower dimensional latent state space enables more efficient planning. The \emph{forward model} is an artificial neural network with a Gated Recurrent Unit (GRU, \citealp{https://doi.org/10.48550/arxiv.1409.0473}) that aggregates temporal information to deal with partial observability. The forward model receives a latent state $z_t$ and an action $a_t$ together with its hidden state $h_t$ from a previous time step $t$ and outputs a next latent state estimate $\hat{z}_{t+1}$, a next reward estimate $\hat{r}_{t}$, and a next hidden state $h_{t+1}$.  The \emph{termination model} $\omega_{\eta}$ receives a latent state $z_t$ and outputs an estimate $\hat{\delta_t} \in [0, 1]$ of whether $z_t$ corresponds to an absorbing state (end of a so-called episode).

The parameters of these three models are iteratively updated in a specific order (autoencoder, forward model, then termination model) using mini-batches stochastic gradient descent. A batch $\mathcal{B} = \{\{ (s_{i,t},a_{i,t},r_{i,t},s_{i,t+1},\delta_{i,t})\}^{L - 1}_{t=0} \}_{i=0}^{B-1}$ is composed of $B$ sequences of length $L$. Each sequence corresponds to an episode chosen at random from a replay buffer $\mathcal{D}$ with capacity $C$ and starts at a time step chosen at random within the corresponding episode.

First, the autoencoder parameters $\chi$ are updated to minimize the reconstruction loss $\mathcal{L}_{\text{AE}}$ given by
\begin{equation} 
    \label{eq:cae}
    \mathcal{L}_{\text{AE}}(\chi) = 
    \frac{1}{BL} \sum_{i=0}^{B-1} \sum_{t = 0}^{L -1}  \| D_{\chi}(E_{\chi}(s_{i,t})) - s_{i,t} \|^2. 
\end{equation}
If we let $z_{i,t} = E_{\chi}(s_{i,t})$, the forward model parameters $\theta$ are then updated to minimize the loss $\mathcal{L}_{\text{F}}$ given by
\small
\begin{align}
    \label{eq:fw}
\mathcal{L}_{\text{F}}(\theta) &= \frac{1}{BL} \sum_{i=0}^{B-1} \sum_{t=0}^{L-1}  \| \hat{z}_{i,t+1} - z_{i, t+1} \|^2  + ( \hat{r}_{i,t} - r_{i, t})^2,
\end{align}
\normalsize
where $(\hat{z}_{i,t+1}, \hat{r}_{i, t}, h_{i, t+1} )= f_\theta(z_{i,t}, a_{i,t}, h_{i, t})$ and $h_{i,0} = 0$ for every $i$. Finally, the termination model parameters $\eta$ are updated to minimize the loss $\mathcal{L}_{\text{T}}$ given by
\begin{align}
    \label{eq:tm}
    \mathcal{L}_{\text{T}}(\eta) &=  \frac{1}{BL} \sum_{i=0}^{B-1} \sum_{t=0}^{L-1} (\omega_{\eta}(z_{i, t+1}) - \delta_{i, t})^2,
\end{align}
where $\delta_{i, t}$ indicates whether $s_{i,t+1}$ is an absorbing state.

\subsection{Uncertainty model}
\label{sec:posterior_distributions}

PSDRL represents its uncertainty about the forward model using the neural-linear method \citep{snoek2015scalable}. This approach achieves remarkable success when combined with posterior sampling in the contextual bandits setting, especially when compared to other (often much more computationally expensive) implementations of Bayesian neural networks \citep{riquelme2018deep}.

\textbf{Linear forward model.} 
In our context, the neural-linear approach models
uncertainty over the parameters $W$ of a forward model $g_W: \mathcal{X} \to \mathcal{Y}$ that ideally maps each possible vector $x = (z, a, h)$ to a vector $y = (z', r')$, where $z$ is a latent state, $a$ is an action, $h$ is a hidden state, $z'$ is the resulting latent state, and $r'$ is the resulting reward.

First, suppose there is a known feature map $\phi : \mathcal{X} \to \mathbb{R}^k$ such that $y = g_{W^*}(x) = W^* \phi(x)$ for every input $x \in \mathcal{X}$ and some 
unknown matrix $W^*$. In words, suppose that the output of the (unknown) forward model $g_{W^*}$ is the result of multiplying an (unknown) parameter matrix $W^*$ by a (known) feature vector $\phi(x)$ that represents the input $x$. Under this assumption, Bayesian linear regression \cite{Rasmussen2004} may be used to model uncertainty over each row $w_j$ of the parameter matrix $W^*$, which is associated with predicting the $j$-th element $y_j$ of the output vector $y$ given the input $x$. Note that $y_j$ is either an element of the next latent state $z'$ or a reward $r'$.

In order to enable efficient inference, PSDRL supposes that the prior density for the vector $w_j$ is given by $\mathcal{N}(w_j \mid 0, \sigma_j^2 I)$, where $\mathcal{N}(\cdot \mid \mu, \Sigma)$ denotes the multivariate normal density function with mean $\mu$ and covariance matrix $\Sigma$ and $I$ denotes the identity matrix. If the index $j$ corresponds to a latent state element, we let $\sigma_j^2 = \sigma_S^2$ for a hyperparameter $\sigma_S^2$. If the index $j$ corresponds to a reward, we let $\sigma_j^2 = \sigma_R^2$ for a hyperparameter $\sigma_R^2$.

Consider a dataset $\{ (x^{(i)}, y^{(i)}) \}_{i=1}^N$ composed of all latent state transitions derived from the replay buffer $\mathcal{D}$. Let $\Phi$ denote a matrix whose $i$-th row is given by $\phi_i = \phi(x^{(i)})$, and let $t_j$ denote the $j$-th vector of targets such that $t_j = (y^{(1)}_j, y^{(2)}_j, \ldots, y^{(N)}_j)$. 
 In order to account for potential modeling errors while still enabling efficient inference, PSDRL also supposes that the likelihood of the parameter vector $w_j$ is given by $\mathcal{N}(t_j \mid \Phi w_j, \sigma^2I)$, where the
noise variance $\sigma^2$ is a hyperparameter.
Under these assumptions, the posterior density for the vector $w_j$ is given by $\mathcal{N}(w_j \mid \mu_j, \Sigma_j)$, where 
\begin{equation}
\label{eq:posterior}
\Sigma_j^{-1} = \sigma_j^{-2} I + \sigma^{-2}\Phi^T\Phi \quad \text{and} \quad
\mu_j = \sigma^{-2} \Sigma_j \Phi^T t_j.
\end{equation}

Let $\tilde{w}_j$ denote a parameter vector drawn from this posterior distribution, and let $\tilde{W}$ denote a matrix whose $j$-th row is given by $\tilde{w}_j$. For a given feature map $\phi$, the forward model $g_{\tilde{W}}$ may be employed to predict the output $\hat{y} = g_{\tilde{W}}(x) = \tilde{W} \phi(x)$ for a given input $x$. In other words,  $g_{\tilde{W}}$ corresponds to an environment model sampled from the posterior, as required by Posterior Sampling for Reinforcement Learning.

\textbf{Feature map learning.} The previous paragraphs have presupposed the existence of a known feature map $\phi : \mathcal{X} \to \mathbb{R}^k$ such that an output $y$ could be predicted by $y = g_{W^*}(x) = W^* \phi(x)$ for every input $x \in \mathcal{X}$ and some unknown matrix $W^*$. In practice, the neural-linear approach derives this feature map from an (iteratively trained) forward model. More concretely, the architecture of an artificial neural network $f_{\theta}$ is chosen such that its prediction $\hat{y}$ for an input $x$ is given by $\hat{y} = f_{\theta}(x) = W\phi_\theta(x)$, where $\phi_\theta$ is a feature map subnetwork and the matrix $W$ is contained in $\theta$. 

In summary, the neural-linear approach represents uncertainty solely over the parameters of the output layer of an artificial neural network whose output layer is a linear function of the last hidden layer while disregarding uncertainty about the parameters of earlier layers.

\subsection{Planning} 
\label{sec:posterior-sampling}

Posterior Sampling for Reinforcement Learning prescribes sampling a forward model $f_{\tilde{\theta}}$ from the corresponding posterior distribution and finding and following an optimal policy $\tilde{\pi}^*$ for this model until the end of an episode. Naturally, it is infeasible to find such an optimal policy in the non-tabular setting. This section describes how PSDRL efficiently searches for a policy for a sampled forward model.

\textbf{Value function approximation.} Consider once again a batch $\mathcal{B}$ composed of $B$ sequences of length $L$ obtained from the replay buffer $\mathcal{D}$. Let $f_{\tilde{\theta}}$ denote a forward model sampled from the posterior such that $\tilde{\theta}$ is composed of a sampled matrix $\tilde{W}$ and parameters of a feature map subnetwork. PSDRL attempts to approximate the value function $V_{\tilde{\pi}^*}$ of an optimal policy $\tilde{\pi}^*$ for the sampled forward model $f_{\tilde{\theta}}$ through a continual procedure that updates the parameters of an artificial neural network $V_{\psi}$. The parameters $\psi$ are updated through mini-batches stochastic gradient descent to minimize the loss $\mathcal{L}_V$ given by
\begin{align}
&
 \mathcal{L}_V(\psi) = \frac{1}{BL} \sum_{i = 0}^{B - 1} \sum_{t = 0}^{L-1} \mathcal{L}_V^{(i,t)}(\psi),
 \label{eq:valueloss}\\
& \text{where} \nonumber \\ 
& \mathcal{L}_V^{(i,t)}(\psi) = \left( V_{\psi}(z_{i, t}, h_{i,t}) - \max_{a} \left[ \hat{r}^{(a)}_{i, t} + \gamma \hat{v}_{i,t+1}^{(a)}\right] \right)^2, \nonumber \\
&(\hat{z}_{i,t+1}^{(a)}, \hat{r}_{i, t}^{(a)}, h_{i, t+1}^{(a)} ) = f_{\tilde{\theta}}(z_{i,t}, a, h_{i, t}), \label{eq:next_reward} \\
&\hat{v}_{i,t+1}^{(a)} = \mathds{1}[\omega_{\eta}(\hat{z}_{i,t+1}^{(a)}) < 0.5]V_{\psi'}(\hat{z}_{i,t+1}^{(a)}, h_{i, t+1}^{(a)} ),
\label{eq:predicted_value}
\end{align}
$z_{i,t} = E_{\chi}(s_{i,t})$, $h_{i,0} = 0$, $h_{i,t + 1} = h_{i,t}^{(a_{i,t})}$, and $\psi'$ are parameters stored from a previous iteration.

This approach is highly related to fitted value iteration \citep{munos2008finite}. In summary, the predicted value of each latent state is updated to become closer to the maximum (considering all actions) predicted next reward plus discounted value predicted for the next predicted latent state (treated as a constant). The current forward model $f_{\tilde{\theta}}$ is responsible for predicting next latent states and rewards.

\textbf{Greedy policy.} PSDRL derives a (greedy) policy $\tilde{\pi}$ from the value network $V_\psi$ such that
\begin{align}
\tilde{\pi}(z_t,h_t) = \argmax_{a} \left[ \hat{r}^{(a)}_{t} + \gamma \hat{v}_{t+1}^{(a)} \right], \label{eq:greedypolicy}
\end{align}
where $z_t$ is a latent state, $h_t$ is a hidden state, $\hat{r}^{(a)}_{t}$ is the predicted reward after action $a$, and $\hat{v}_{t+1}^{(a)}$ is the predicted value for the predicted next latent state, which are obtained in analogy with Equations \ref{eq:next_reward} and \ref{eq:predicted_value} (except $\psi$ overrides $\psi'$).

Algorithm \ref{PSDRL} summarizes PSDRL (forward models are sampled every $m$ time steps instead of episodically).

\begin{algorithm}[ht]
\caption{PSDRL}\label{PSDRL}
\begin{algorithmic}[1]
\STATE $\mathcal{D} \gets $ empty (FIFO) buffer with capacity $C$
\STATE $s_0 \gets $ initial state
\STATE $h_0 \gets 0$
\FOR{each $t \in \{0, \ldots, T -1\}$}
\IF{$t \mod m = 0$}
\STATE $\chi \gets $ update for autoencoder loss $\mathcal{L}_{\text{AE}}$ (Eq. \ref{eq:cae})
\STATE $\theta \gets $ update for forward model loss $\mathcal{L}_{\text{F}}$ (Eq. \ref{eq:fw})
\STATE $\eta \gets $ update for termination model loss $\mathcal{L}_{\text{T}}$ (Eq. \ref{eq:tm})
\STATE $\mu_j, \Sigma_j \gets $ update posterior for each $j$ (Eq. \ref{eq:posterior}) 
\STATE $\tilde{W} \gets $ sample from posteriors 
\STATE $ f_{\tilde{\theta}} \gets $ forward model derived from $\tilde{W}$ and feature map subnetwork $\phi_{\theta}$
\STATE $\psi \gets $ update for value network loss $\mathcal{L}_V$ (Eq. \ref{eq:valueloss}) based on the sampled forward model $f_{\tilde{\theta}}$
\ENDIF
\STATE $z_t \gets E_\chi(s_t)$
\STATE $a_t \gets \tilde{\pi}(z_t, h_t)$ (Eq. \ref{eq:greedypolicy})
\STATE $r_t,s_{t+1},\delta \gets$ outcome of action $a_t$
\STATE $h_{t+1} \gets$ corresponding output from $f_{\tilde{\theta}}(z_t, a_t, h_{t})$
\STATE $\mathcal{D} \gets \mathcal{D} \cup \{(s_{t},a_{t},r_{t},s_{t+1},\delta)\}$ 
\IF{$\delta = 1$}
    \STATE $s_{t+1} \gets $ initial state
    \STATE $h_{t + 1} \gets $ 0
\ENDIF
\ENDFOR
\end{algorithmic}
\label{value}
\end{algorithm}

\subsection{Rationale}
\label{sec:justification}

This section discusses crucial implementation choices that enable PSDRL to scale up to complex environments. 

The choice of a latent state space transition model (Sec. \ref{sec:dynamics_model}) enables more efficient planning in comparison with employing a (higher-dimensional) state space transition model. It is also justified by the recent success of latent state space models in model-based reinforcement learning 
. Similarly, 
the choice of the neural-linear approach for modeling uncertainty over forward models (Sec. \ref{sec:posterior_distributions}) enables (relatively) inexpensive posterior sampling. Furthermore, the neural-linear approach has achieved remarkable empirical success 
with posterior sampling in the contextual bandits setting.

However, the choice of planning algorithm (Sec.  \ref{sec:posterior-sampling}) is comparatively much more involved since it requires careful consideration of the strengths and weaknesses of the aforementioned components. In particular, the planning algorithm needs to avoid the following potential pitfalls.

\textbf{Recency bias.} The posterior update step (Alg. \ref{PSDRL}, line 9) implicitly requires recomputing all latent state transitions derived from the replay buffer $\mathcal{D}$. This is a consequence of the fact that the latent state representations are learned (as opposed to given), which precludes the typical Bayesian approach of using a posterior as the new prior when additional data is acquired. Therefore, although the forward model parameters $\theta$ are potentially influenced by all previous transitions (since they reflect all previous updates), the posterior over (output layer) parameters is strictly influenced by the transitions in the replay buffer. Because the replay buffer is necessarily limited in capacity, this introduces a bias that may severely impact the efficiency of the algorithm. 

Previous works that adapt Posterior Sampling for Reinforcement Learning to non-tabular settings \citep{tziortziotis2013linear, fan2021modelbased} sidestep this pitfall by employing fixed state representations or an unlimited replay buffer, which pose significant scalability challenges. 

PSDRL addresses this recency bias by retaining information obtained from previous sampled models in a value network $V_\psi$ through continual training. More concretely, the value network trained for a previous sampled model is used as the starting point for the current sampled model. Therefore, even if the entire replay buffer $\mathcal{D}$ is composed of relatively uninformative transitions (for instance, zero-reward transitions), the decisions of the agent are potentially influenced by previous planning results. This is notably distinct from the planning approaches that rely entirely on a current sampled model employed in previous works. In comparison with search methods, derivative-free optimization, and policy gradient methods, value-function approximation is naturally suited to aggregate information across sampled models, which justifies its choice. Importantly, this continual approach also improves planning efficiency when the current sampled model is similar to the previous sampled model. However, training a value function approximator across sampled models introduces another potential pitfall.

\textbf{Status quo bias.} Posterior Sampling for Reinforcement Learning prescribes sampling a forward model from the corresponding posterior distribution and following an optimal policy for this model for at least one episode. The fact that a posterior sampling agent may radically change its behavior based on a sampled model is crucial for efficient exploration. Therefore, employing a value network that aggregates information across sampled models as a starting point to train a value network for a current model may bias the agent in a way that is detrimental to exploration. 

PSDRL addresses this status quo bias by choosing actions that maximize the one-step return, which combines the next reward predicted by the sampled model with the value predicted by the value network for the next (latent) state predicted by the sampled model (Eq. \ref{eq:greedypolicy}). For instance, instead of considering the one-step return, it would be possible to train a typical action-value network $Q_\psi$ and choose the action $\argmax_{a} Q_\psi(z_t, h_t, a)$ for any given latent state $z_t$ and hidden state $z_t$. However, unless $Q_\psi$ changes significantly after a relatively brief training procedure (Alg. \ref{PSDRL}, line 12), the resulting policy (Alg. \ref{PSDRL}, line 15) could fail to incorporate sufficient feedback from the current sampled model. In order to further reduce the status quo bias, it would also be possible (but more expensive) to consider the $k$-step return of the greedy policy with respect to $V_\psi$ (or $Q_\psi$).

In summary, there is a natural trade-off between recency bias and status quo bias. Section \ref{sec:ablation-studies} reports an ablation study that conclusively shows that disregarding planning results from previous sampled models is not a suitable alternative under realistic computational constraints, whereas our choices lead to efficient exploration.

\begin{table}[t]
\caption{Median and mean human-normalized score
\citep{Mnih2015-dx}, mean record-normalized score 
\citep{recordmean}, and mean clipped record-normalized score
\citep{Hafner2020-ry}.}
\label{atari-scores}
\vskip 0.15in
\begin{center}
\footnotesize
\begin{tabular}{lcccc}
\toprule
\makecell[l]{\textbf{Metric}} 
& \makecell[c]{\textbf{PSDRL}} 
& \makecell[c]{\textbf{Dv2}} 
& \makecell[c]{\textbf{B+P}} 
& \makecell[c]{\textbf{SU}} \\
\midrule
Gamer Median & \textbf{23}\% & 18\% & 7\% & 0\%\\
Gamer Mean & 58\% & \textbf{100}\% & 0\% & 21\%\\
Record Mean & 6\% &\textbf{ 8}\% & 0\% & 0\%\\
Clipped Record Mean &6\% & \textbf{8}\% & 0\% & 0\%\\
\bottomrule
\end{tabular}
\end{center}
\vskip -0.1in
\end{table}

\section{Experiments}
\label{sec:experiments}


\subsection{Experimental protocol}

\textbf{Environments.} We provide an experimental comparison between PSDRL and other algorithms on 55 Atari 2600 games that are commonly used in the literature \citep{Mnih2015-dx}. We evaluate each algorithm on all 55 games using no full action space, no access to life information, no sticky actions, and an action repeat of four for three random seeds. In order to keep to a feasible computational budget, we restrict environment steps to 1M (4M frames).

\textbf{Algorithms.} Since previous model-based posterior sampling algorithms do not scale to complex environments, we make a comparison with two state-of-the-art randomized value function (RVF) algorithms: Bootstrapped DQN with randomized priors (B+P, \citealp{osband2018randomized}) and Successor Uncertainties (SU, \citealp{janz2019successor}). Additionally, we make a comparison with the model-based algorithm DreamerV2 (Dv2, \citealp{Hafner2020-ry}), which is the state-of-the-art single-GPU algorithm for the Atari benchmark. 

\textbf{Evaluation metrics.} Traditionally, authors report the mean and median human-normalized scores across all games (gamer mean and gamer median) \citep{Mnih2015-dx}. A human-normalized score of zero corresponds to a randomly acting agent, whereas a human-normalized score of one corresponds to a professional gamer.
However, \citet{recordmean} convincingly argue that the gamer mean can be misleading due to outliers, whereas the gamer median can be misleading if almost half of the scores are zero. Therefore, they propose to normalize according to the human world records instead. \citet{Hafner2020-ry} further propose the clipped record-normalized score. A record-normalized score of zero corresponds to a randomly acting agent, whereas a record-normalized score of one corresponds to a world record performance. Performances that exceed the world record receive a clipped record-normalized score of one. We report both traditional human-normalized scores and (clipped) record-normalized scores, which are computed using the average evaluation return across 1M environment steps, measured every 10k environment steps.

\textbf{Hyperparameters.} As the original hyperparameters for the baseline algorithms were tuned for 200M frames, and sticky actions in the case of Dv2, we have tuned each baseline for the deterministic 4M frames setting to guarantee a fair comparison. Hyperparameters are obtained with an exhaustive grid search on six Atari games commonly used for this purpose \citep{munos2016safe, janz2019successor}: \textsc{Asterix}, \textsc{Enduro}, \textsc{Freeway}, \textsc{Hero}, \textsc{Qbert}, and \textsc{Seaquest}.

\textbf{Reproducibility.} The source code for replicating all experiments is available as supplementary material. A description of the hyperparameter search procedure and other implementation decisions are available in Appendix \ref{app:hyperparams}.

\subsection{Main results}
\label{sec:main-results}
\begin{figure*}[h!]
\vskip 0.2in
\begin{center}
    \includegraphics[width=0.85\linewidth]{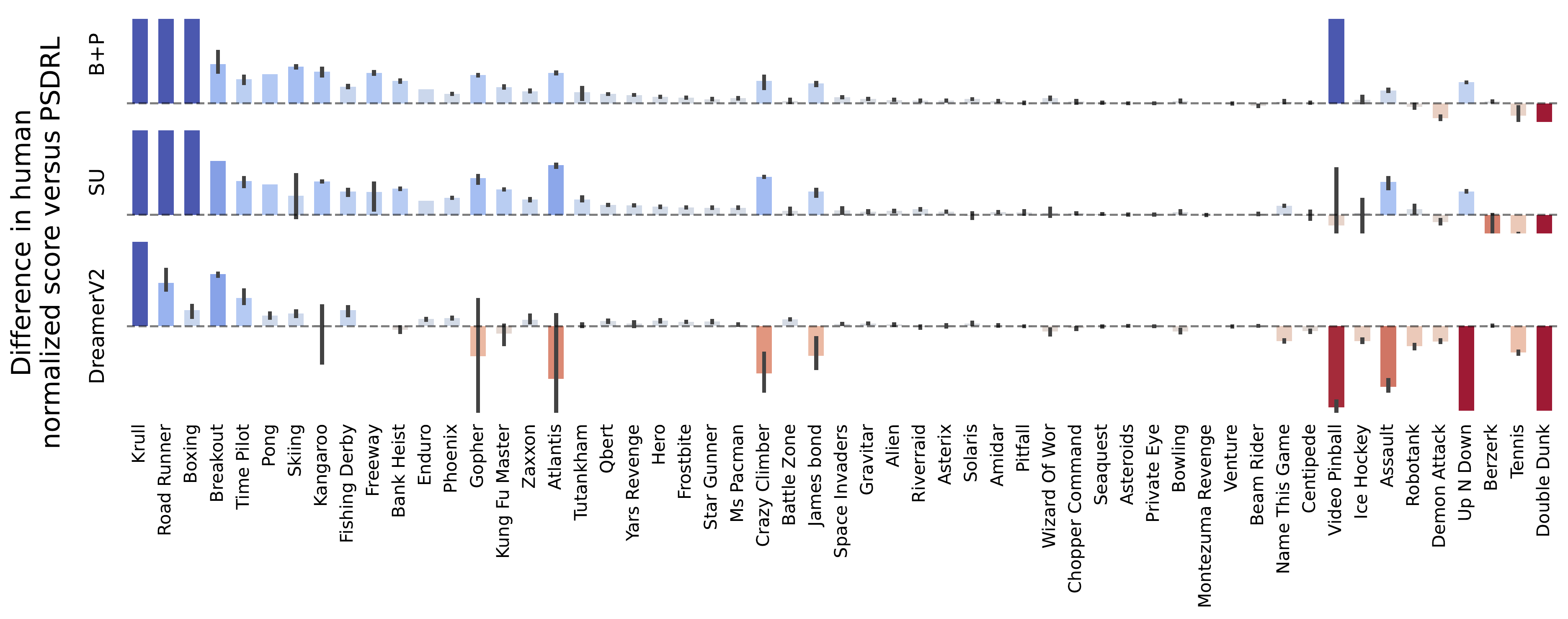}
    \caption{A comparison between PSDRL and B+P (top), SU (middle), and Dv2 (bottom) for all 55 Atari games. Blue indicates that PSDRL scores higher than the baseline, red indicates the opposite. Y-axis has been clipped to $[-2.5, 2.5]$ to facilitate visualization.}
    \label{fig:all_scores}
\end{center}
\vskip -0.1in
\end{figure*}

The results for each evaluation metric are summarized in Table \ref{atari-scores}. Note that although SU significantly outperforms B+P in terms of gamer mean, this is almost entirely due to its excellent performance on \textsc{Berzerk}, which is just one example of how this traditional metric is flawed. A more informative visualization comparing PSDRL to each baseline algorithm in terms of human-normalized score is presented in Figure \ref{fig:all_scores} (Appendix \ref{sec:normalized_visuals} details this visualization). The raw game scores and corresponding learning curves for each individual game can be found in Appendices \ref{app:numericalscores} and \ref{app:learningcurves}, respectively. Remarkably, the results show that PSDRL is competitive with the state-of-the-art algorithm Dv2 in terms of sample efficiency while using a similar computational budget. Additionally, PSDRL substantially outperforms the state-of-the-art RVF baselines in almost all of the games. 

Appendix \ref{app:reconstructions} shows that PSDRL is able to decode latent state predictions obtained from sampled forward models close to perfectly for many games at the end of training, even when the corresponding state images have many details. Exceptionally, the fact that PSDRL and Dv2 receive downscaled $64 \times 64$ images (compared to $84 \times 84$ for B+P) as state representations makes it difficult for their agents to distinguish between the enemies depicted with small differences in color and shape in games like \textsc{BeamRider} within 1M environment steps, which explains why they are outperformed by B+P in this case (see Table \ref{app:numericalscores}). Therefore, the state representation and the model capacity of an autoencoder introduce an additional trade-off between sample efficiency and computational efficiency. 

\begin{figure}[t]
\vskip 0.2in
\begin{center}
    \includegraphics[width=\linewidth]{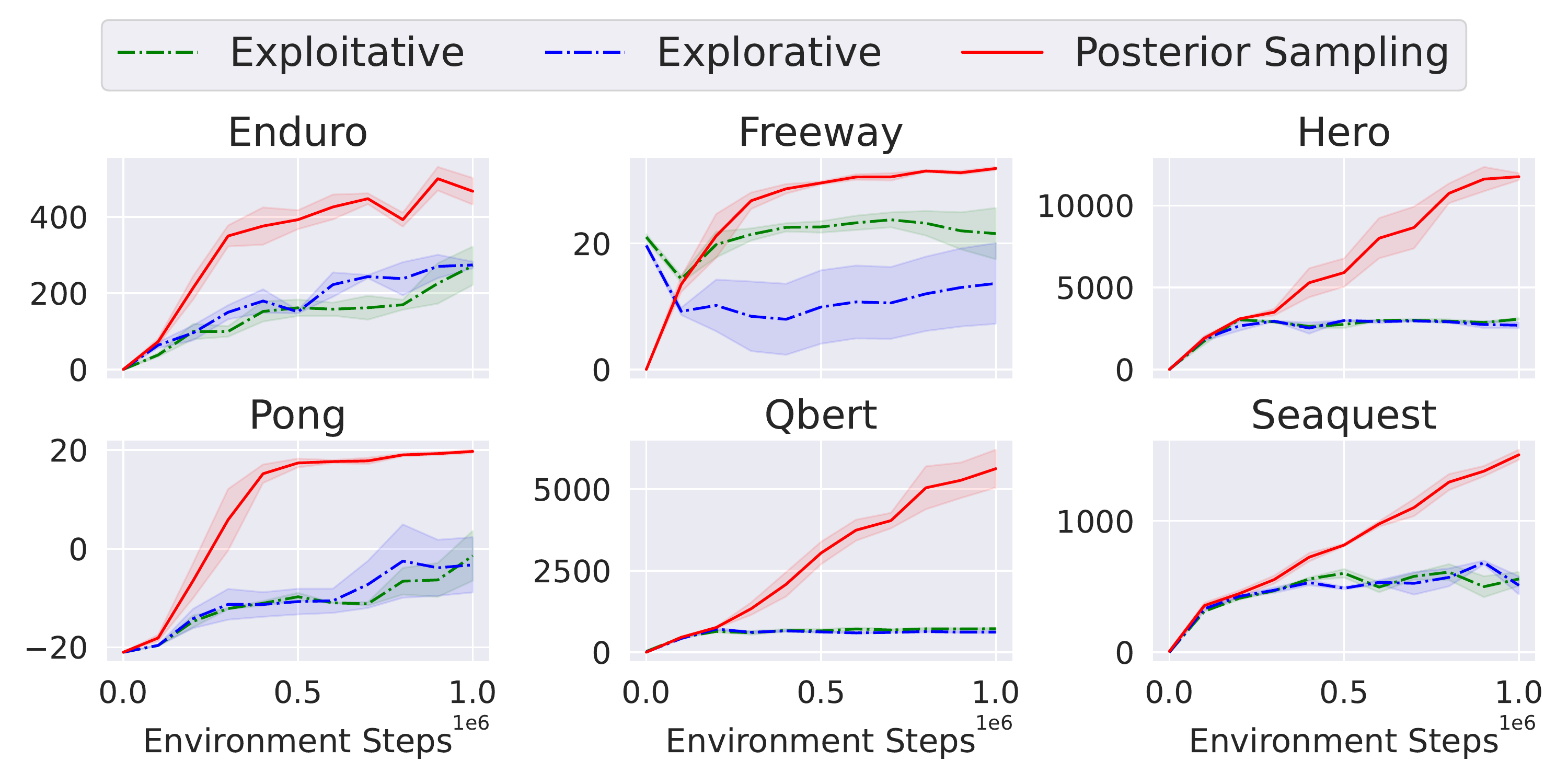}
    \caption{Average evaluation episode returns comparing posterior sampling to \textit{exploitative} and \textit{explorative} $\epsilon$-greedy approaches.}
    \label{fig:exploration_ablation}
\end{center}
\vskip -0.1in
\end{figure}

\subsection{Ablation studies} 
\label{sec:ablation-studies}

This section details two ablation studies that we conducted to better understand the importance of posterior sampling and our continual value function approximation approach.

\begin{figure}[b]
\vskip 0.2in
\begin{center}
    \includegraphics[width=\linewidth]{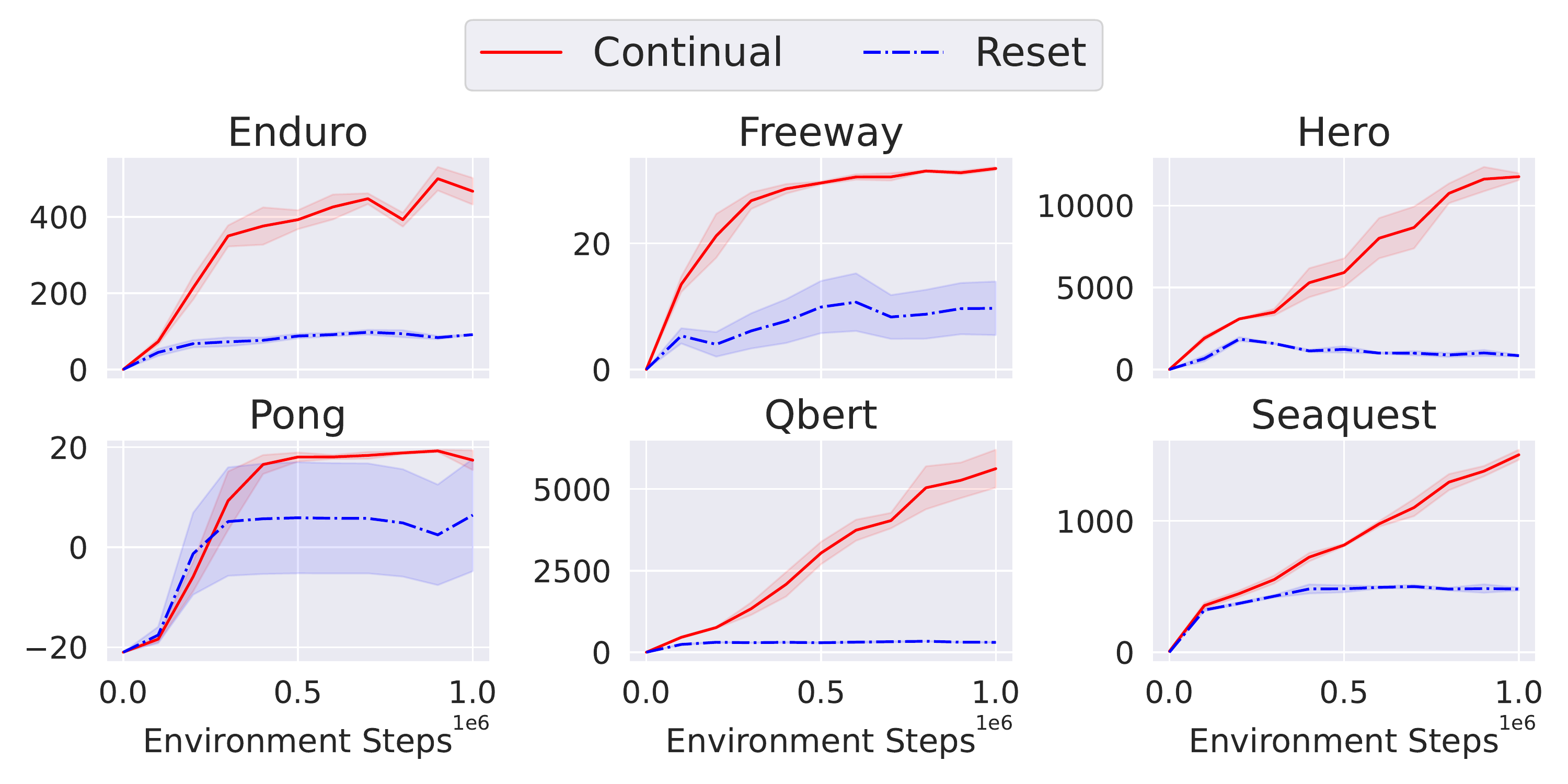}
    \caption{Average evaluation episode returns comparing planning continually with planning with newly initialized value functions.}
    \label{fig:value_ablation}
\end{center}
\vskip -0.1in
\end{figure}

\textbf{Posterior sampling.} In order to investigate the significance of posterior sampling in PSDRL, we make a comparison with two $\epsilon$-greedy approaches that select exploratory actions at random with probability $\epsilon$. Concretely, these approaches employ the same architecture and hyperparameters as PSDRL but do not sample forward models from the corresponding posterior. Instead, these approaches always use the current forward model parameters $\theta$ to make predictions. In order to provide a fair comparison, we linearly anneal $\epsilon$ from 1 to 0.01 in two different schemes: an \textit{exploitative} annealing scheme across 50k environment steps, and an \textit{explorative} annealing scheme across 1M environment steps. The results in Figure \ref{fig:exploration_ablation} clearly show that the success of PSDRL is heavily dependent on the natural balancing between exploration and exploitation provided by posterior sampling. For instance, in the sparse-reward environment \textsc{Freeway}, posterior sampling allows the agent to learn the optimal policy rapidly, whereas the explorative agent is slowed down significantly and the exploitative agent settles on a sub-optimal policy.

\textbf{Continual value network.}
Section \ref{sec:justification} provides the rationale behind our continual approach that trains a value function approximator across sampled models instead of using a newly initialized value network for each sampled model. In order to provide a fair comparison between these two approaches, we quadruple the number of iterations used to train newly initialized value functions at every planning step (Alg \ref{PSDRL}, line 12). The results in Figure \ref{fig:value_ablation} show that, even when the total computational cost is increased by approximately $60\%$, using a newly initialized value network at every planning step is detrimental to sample efficiency. This clearly justifies biasing planning towards the optimal policy for previously sampled models.

\section{Conclusion}
\label{sec:conclusion}

We introduced Posterior Sampling for Deep Reinforcement Learning (PSDRL), the first truly scalable approximation of Posterior Sampling for Reinforcement Learning that retains its model-based essence. Although there are many other promising approaches towards efficient exploration, Posterior Sampling for Reinforcement Learning is the simplest among the potentially scalable and principled methods that is capable of leveraging the strengths of both Bayesian methods and model-based reinforcement learning.

In addition to an efficient architecture for representing uncertainty over latent state space transition models inspired by recent successes in model-based reinforcement learning and contextual multi-armed bandits, our technical contributions include a specially tailored continual planning algorithm based on value-function approximation that addresses two newly identified potential pitfalls in scaling up Posterior Sampling for Reinforcement Learning (recency bias and status quo bias). Our extensive experiments on the Atari benchmark show that PSDRL significantly outperforms previous state-of-the-art randomized value function approaches, its natural model-free counterparts, while being competitive with a state-of-the-art (model-based) reinforcement learning method in both sample efficiency and computational efficiency. Our approach is further validated by ablation studies that show the importance of retaining planning information across sampled models and the superiority of posterior sampling over a naive exploration method.

Similarly to most deep reinforcement learning algorithms, the main weakness of PSDRL is the cost of successfully dealing with traditionally \emph{hard} environments (such as those with very sparse rewards), which are defined by having high visitation complexity or estimation complexity \cite{conserva2022hardness}. Such environments naturally require large replay buffers (to counteract the recency bias), lower forward model sampling frequency (to encourage deep exploration \citep{osband2019deep}), and longer planning times (to obtain better policies). Some of these issues may be mitigated by employing smart replay buffers and strategies such as prioritized experience replay \citep{schaul2016}. Employing a latent state space model architecture that allows incremental posterior updates is another promising approach, although it requires finding a careful balance between reliable uncertainty quantification and computational efficiency.

There are numerous possibilities for future work. The most important and demanding is to investigate the many feasible combinations of latent state space model architectures with Bayesian neural networks. Our experience indicates that it is crucial to tailor the corresponding planning algorithm while considering the strengths and weaknesses of the remaining components. Extending our approach to deal with more general non-deterministic environments is also especially important. Finally, significant insight into our approach may be gained from studying it from a theoretical perspective.

\section*{Acknowledgements}
This research was financially supported by the Intelligent Games and Games Intelligence CDT (IGGI;EP/S022325/1) and utilized Queen Mary's Apocrita HPC facility (\url{http://doi.org/10.5281/zenodo.438045}).

\bibliography{main}

\begin{thebibliography}{55}
\providecommand{\natexlab}[1]{#1}
\providecommand{\url}[1]{\texttt{#1}}
\expandafter\ifx\csname urlstyle\endcsname\relax
  \providecommand{\doi}[1]{doi: #1}\else
  \providecommand{\doi}{doi: \begingroup \urlstyle{rm}\Url}\fi

\bibitem[Agrawal \& Jia(2017)Agrawal and Jia]{agrawal2017optimistic}
Agrawal, S. and Jia, R.
\newblock
  \href{https://papers.nips.cc/paper/2017/hash/3621f1454cacf995530ea53652ddf8fb-Abstract.html}{Optimistic
  posterior sampling for reinforcement learning: worst-case regret bounds}.
\newblock \emph{Advances in Neural Information Processing Systems}, 30, 2017.

\bibitem[Aytar et~al.(2018)Aytar, Pfaff, Budden, Paine, Wang, and
  De~Freitas]{aytar2018playing}
Aytar, Y., Pfaff, T., Budden, D., Paine, T., Wang, Z., and De~Freitas, N.
\newblock
  \href{https://papers.nips.cc/paper/2018/hash/35309226eb45ec366ca86a4329a2b7c3-Abstract.html}{Playing
  hard exploration games by watching youtube}.
\newblock \emph{Advances in Neural Information Processing Systems}, 31, 2018.

\bibitem[Azar et~al.(2017)Azar, Osband, and Munos]{azar2017minimax}
Azar, M.~G., Osband, I., and Munos, R.
\newblock \href{https://proceedings.mlr.press/v70/azar17a.html}{Minimax regret
  bounds for reinforcement learning}.
\newblock In \emph{International Conference on Machine Learning}, volume~70,
  pp.\  263--272. PMLR, 2017.

\bibitem[Azizzadenesheli et~al.(2018)Azizzadenesheli, Brunskill, and
  Anandkumar]{8503252}
Azizzadenesheli, K., Brunskill, E., and Anandkumar, A.
\newblock \href{https://ieeexplore.ieee.org/document/8503252}{Efficient
  Exploration Through Bayesian Deep Q-Networks}.
\newblock In \emph{2018 Information Theory and Applications Workshop (ITA)},
  pp.\  1--9, 2018.

\bibitem[Badia et~al.(2019)Badia, Sprechmann, Vitvitskyi, Guo, Piot,
  Kapturowski, Tieleman, Arjovsky, Pritzel, Bolt, et~al.]{badia2019never}
Badia, A.~P., Sprechmann, P., Vitvitskyi, A., Guo, D., Piot, B., Kapturowski,
  S., Tieleman, O., Arjovsky, M., Pritzel, A., Bolt, A., et~al.
\newblock \href{https://openreview.net/forum?id=Sye57xStvB}{Never Give Up:
  Learning Directed Exploration Strategies}.
\newblock In \emph{International Conference on Learning Representations}, 2019.

\bibitem[Badia et~al.(2020)Badia, Piot, Kapturowski, Sprechmann, Vitvitskyi,
  Guo, and Blundell]{badia2020agent57}
Badia, A.~P., Piot, B., Kapturowski, S., Sprechmann, P., Vitvitskyi, A., Guo,
  Z.~D., and Blundell, C.
\newblock \href{https://proceedings.mlr.press/v119/badia20a.html}{Agent57:
  Outperforming the atari human benchmark}.
\newblock In \emph{International Conference on Machine Learning}, pp.\
  507--517. PMLR, 2020.

\bibitem[Bahdanau et~al.(2015)Bahdanau, Cho, and
  Bengio]{https://doi.org/10.48550/arxiv.1409.0473}
Bahdanau, D., Cho, K.~H., and Bengio, Y.
\newblock \href{http://arxiv.org/abs/1409.0473}{Neural machine translation by
  jointly learning to align and translate}.
\newblock In \emph{International Conference on Learning Representations}, 2015.

\bibitem[Bai et~al.(2021)Bai, Wang, Han, Hao, Garg, Liu, and
  Wang]{bai2021principled}
Bai, C., Wang, L., Han, L., Hao, J., Garg, A., Liu, P., and Wang, Z.
\newblock \href{https://proceedings.mlr.press/v139/bai21d.html}{Principled
  exploration via optimistic bootstrapping and backward induction}.
\newblock In \emph{International Conference on Machine Learning}, pp.\
  577--587. PMLR, 2021.

\bibitem[Bellemare et~al.(2016)Bellemare, Srinivasan, Ostrovski, Schaul,
  Saxton, and Munos]{bellemare2016unifying}
Bellemare, M., Srinivasan, S., Ostrovski, G., Schaul, T., Saxton, D., and
  Munos, R.
\newblock
  \href{https://papers.nips.cc/paper/2016/hash/afda332245e2af431fb7b672a68b659d-Abstract.html}{Unifying
  count-based exploration and intrinsic motivation}.
\newblock \emph{Advances in Neural Information Processing Systems}, 29, 2016.

\bibitem[Bellemare et~al.(2013)Bellemare, Naddaf, Veness, and
  Bowling]{bellemare2013arcade}
Bellemare, M.~G., Naddaf, Y., Veness, J., and Bowling, M.
\newblock \href{https://jair.org/index.php/jair/article/view/10819}{The arcade
  learning environment: An evaluation platform for general agents}.
\newblock \emph{Journal of Artificial Intelligence Research}, 47:\penalty0
  253--279, 2013.

\bibitem[Berner et~al.(2019)Berner, Brockman, Chan, Cheung, Debiak, Dennison,
  Farhi, Fischer, Hashme, Hesse, et~al.]{berner2019dota}
Berner, C., Brockman, G., Chan, B., Cheung, V., Debiak, P., Dennison, C.,
  Farhi, D., Fischer, Q., Hashme, S., Hesse, C., et~al.
\newblock \href{https://arxiv.org/abs/1912.06680}{Dota 2 with large scale deep
  reinforcement learning}.
\newblock \emph{arXiv preprint}, 2019.

\bibitem[Botev et~al.(2013)Botev, Kroese, Rubinstein, and
  L’Ecuyer]{BOTEV201335}
Botev, Z.~I., Kroese, D.~P., Rubinstein, R.~Y., and L’Ecuyer, P.
\newblock
  \href{https://www.sciencedirect.com/science/article/abs/pii/B9780444538598000035}{Chapter
  3 - The Cross-Entropy Method for Optimization}.
\newblock In \emph{Handbook of Statistics}, volume~31, pp.\  35--59. Elsevier,
  2013.

\bibitem[Camacho \& Bordons(2007)Camacho and Bordons]{24286}
Camacho, E.~F. and Bordons, C.
\newblock
  \emph{\href{https://link.springer.com/book/10.1007/978-0-85729-398-5}{Model
  Predictive control}}.
\newblock Advanced Textbooks in Control and Signal Processing. Springer London,
  2007.

\bibitem[Chen et~al.(2017)Chen, Sidor, Abbeel, and Schulman]{chen2017ucb}
Chen, R.~Y., Sidor, S., Abbeel, P., and Schulman, J.
\newblock \href{https://arxiv.org/abs/1706.01502}{Ucb exploration via
  q-ensembles}.
\newblock \emph{arXiv preprint arXiv:1706.01502}, 2017.

\bibitem[Conserva \& Rauber(2022)Conserva and Rauber]{conserva2022hardness}
Conserva, M. and Rauber, P.
\newblock Hardness in markov decision processes: Theory and practice.
\newblock In \emph{Advances in Neural Information Processing Systems},
  volume~35, 2022.

\bibitem[Ecoffet et~al.(2021)Ecoffet, Huizinga, Lehman, Stanley, and
  Clune]{ecoffet2021first}
Ecoffet, A., Huizinga, J., Lehman, J., Stanley, K.~O., and Clune, J.
\newblock \href{https://www.nature.com/articles/s41586-020-03157-9}{First
  return, then explore}.
\newblock \emph{Nature}, 590\penalty0 (7847):\penalty0 580--586, 2021.

\bibitem[Engel et~al.(2003)Engel, Mannor, and Meir]{engel2003bayes}
Engel, Y., Mannor, S., and Meir, R.
\newblock \href{https://www.aaai.org/Library/ICML/2003/icml03-023.php}{Bayes
  meets Bellman: The Gaussian process approach to temporal difference
  learning}.
\newblock In \emph{International Conference on Machine Learning}, pp.\
  154--161. PMLR, 2003.

\bibitem[Engel et~al.(2005)Engel, Mannor, and Meir]{engel2005reinforcement}
Engel, Y., Mannor, S., and Meir, R.
\newblock
  \href{https://icml.cc/Conferences/2005/proceedings/papers/026_Reinforcement_EngelEtAl.pdf}{Reinforcement
  learning with Gaussian processes}.
\newblock In \emph{International Conference on Machine Learning}, pp.\
  201--208. PMLR, 2005.

\bibitem[Eysenbach et~al.(2018)Eysenbach, Gupta, Ibarz, and
  Levine]{eysenbach2018diversity}
Eysenbach, B., Gupta, A., Ibarz, J., and Levine, S.
\newblock \href{https://openreview.net/forum?id=SJx63jRqFm}{Diversity is All
  You Need: Learning Skills without a Reward Function}.
\newblock In \emph{International Conference on Learning Representations}, 2018.

\bibitem[Fan \& Ming(2021)Fan and Ming]{fan2021modelbased}
Fan, Y. and Ming, Y.
\newblock \href{https://proceedings.mlr.press/v139/fan21b.html}{Model-based
  Reinforcement Learning for Continuous Control with Posterior Sampling}.
\newblock In \emph{International Conference on Machine Learning}, pp.\
  3078--3087. PMLR, 2021.

\bibitem[Flennerhag et~al.(2020)Flennerhag, Wang, Sprechmann, Visin, Galashov,
  Kapturowski, Borsa, Heess, Barreto, and Pascanu]{flennerhag2020temporal}
Flennerhag, S., Wang, J.~X., Sprechmann, P., Visin, F., Galashov, A.,
  Kapturowski, S., Borsa, D.~L., Heess, N., Barreto, A., and Pascanu, R.
\newblock \href{https://arxiv.org/abs/2010.02255}{Temporal difference
  uncertainties as a signal for exploration}.
\newblock \emph{arXiv preprint arXiv:2010.02255}, 2020.

\bibitem[Ha \& Schmidhuber(2018)Ha and Schmidhuber]{Ha2018-jd}
Ha, D. and Schmidhuber, J.
\newblock
  \href{https://papers.nips.cc/paper/2018/hash/2de5d16682c3c35007e4e92982f1a2ba-Abstract.html}{Recurrent
  world models facilitate policy evolution}.
\newblock \emph{Advances in Neural Information Processing Systems}, 31, 2018.

\bibitem[Hafner et~al.(2019{\natexlab{a}})Hafner, Lillicrap, Ba, and
  Norouzi]{hafner2020dream}
Hafner, D., Lillicrap, T., Ba, J., and Norouzi, M.
\newblock \href{https://openreview.net/forum?id=S1lOTC4tDS}{Dream to Control:
  Learning Behaviors by Latent Imagination}.
\newblock In \emph{International Conference on Learning Representations},
  2019{\natexlab{a}}.

\bibitem[Hafner et~al.(2019{\natexlab{b}})Hafner, Lillicrap, Fischer, Villegas,
  Ha, Lee, and Davidson]{hafner2019learning}
Hafner, D., Lillicrap, T., Fischer, I., Villegas, R., Ha, D., Lee, H., and
  Davidson, J.
\newblock \href{https://proceedings.mlr.press/v97/hafner19a.html}{Learning
  latent dynamics for planning from pixels}.
\newblock In \emph{International Conference on Machine Learning}, pp.\
  2555--2565. PMLR, 2019{\natexlab{b}}.

\bibitem[Hafner et~al.(2020)Hafner, Lillicrap, Norouzi, and Ba]{Hafner2020-ry}
Hafner, D., Lillicrap, T.~P., Norouzi, M., and Ba, J.
\newblock \href{https://openreview.net/forum?id=0oabwyZbOu}{Mastering Atari
  with Discrete World Models}.
\newblock In \emph{International Conference on Learning Representations}, 2020.

\bibitem[Janner et~al.(2019)Janner, Fu, Zhang, and Levine]{janner2019trust}
Janner, M., Fu, J., Zhang, M., and Levine, S.
\newblock
  \href{https://papers.nips.cc/paper/2019/hash/5faf461eff3099671ad63c6f3f094f7f-Abstract.html}{When
  to trust your model: Model-based policy optimization}.
\newblock \emph{Advances in Neural Information Processing Systems}, 32, 2019.

\bibitem[Janz et~al.(2019)Janz, Hron, Mazur, Hofmann, Hern{\'a}ndez-Lobato, and
  Tschiatschek]{janz2019successor}
Janz, D., Hron, J., Mazur, P., Hofmann, K., Hern{\'a}ndez-Lobato, J.~M., and
  Tschiatschek, S.
\newblock
  \href{https://proceedings.neurips.cc/paper/2019/hash/1b113258af3968aaf3969ca67e744ff8-Abstract.html}{Successor
  uncertainties: exploration and uncertainty in temporal difference learning}.
\newblock \emph{Advances in Neural Information Processing Systems}, 32, 2019.

\bibitem[Kaiser et~al.(2019)Kaiser, Babaeizadeh, Mi{\l}os, Osi{\'n}ski,
  Campbell, Czechowski, Erhan, Finn, Kozakowski, Levine, et~al.]{Kaiser2019-ts}
Kaiser, {\L}., Babaeizadeh, M., Mi{\l}os, P., Osi{\'n}ski, B., Campbell, R.~H.,
  Czechowski, K., Erhan, D., Finn, C., Kozakowski, P., Levine, S., et~al.
\newblock \href{https://openreview.net/forum?id=S1xCPJHtDB}{Model Based
  Reinforcement Learning for Atari}.
\newblock In \emph{International Conference on Learning Representations}, 2019.

\bibitem[Kingma \& Ba(2015)Kingma and Ba]{kingma2014adam}
Kingma, D.~P. and Ba, J.
\newblock \href{http://arxiv.org/abs/1412.6980}{Adam: A Method for Stochastic
  Optimization}.
\newblock In \emph{International Conference on Learning Representations}, 2015.

\bibitem[Kumar et~al.(2019)Kumar, Fu, Soh, Tucker, and
  Levine]{NEURIPS2019_c2073ffa}
Kumar, A., Fu, J., Soh, M., Tucker, G., and Levine, S.
\newblock
  \href{https://proceedings.neurips.cc/paper/2019/hash/c2073ffa77b5357a498057413bb09d3a-Abstract.html}{Stabilizing
  Off-Policy Q-Learning via Bootstrapping Error Reduction}.
\newblock In \emph{Advances in Neural Information Processing Systems},
  volume~32, 2019.

\bibitem[Kumar et~al.(2020)Kumar, Gupta, and Levine]{kumar2020discor}
Kumar, A., Gupta, A., and Levine, S.
\newblock
  \href{https://proceedings.neurips.cc/paper/2020/hash/d7f426ccbc6db7e235c57958c21c5dfa-Abstract.html}{Discor:
  Corrective feedback in reinforcement learning via distribution correction}.
\newblock \emph{Advances in Neural Information Processing Systems}, 33, 2020.

\bibitem[Masci et~al.(2011)Masci, Meier, Cire{\c{s}}an, and
  Schmidhuber]{masci2011stacked}
Masci, J., Meier, U., Cire{\c{s}}an, D., and Schmidhuber, J.
\newblock
  \href{https://link.springer.com/chapter/10.1007/978-3-642-21735-7_7}{Stacked
  convolutional auto-encoders for hierarchical feature extraction}.
\newblock In \emph{International conference on artificial neural networks},
  pp.\  52--59. Springer, 2011.

\bibitem[M{\'e}nard et~al.(2021)M{\'e}nard, Domingues, Shang, and
  Valko]{menard2021ucb}
M{\'e}nard, P., Domingues, O.~D., Shang, X., and Valko, M.
\newblock \href{http://proceedings.mlr.press/v139/menard21b.html}{UCB Momentum
  Q-learning: Correcting the bias without forgetting}.
\newblock In \emph{International Conference on Machine Learning}, pp.\
  7609--7618. PMLR, 2021.

\bibitem[Mnih et~al.(2015)Mnih, Kavukcuoglu, Silver, Rusu, Veness, Bellemare,
  Graves, Riedmiller, Fidjeland, Ostrovski, et~al.]{Mnih2015-dx}
Mnih, V., Kavukcuoglu, K., Silver, D., Rusu, A.~A., Veness, J., Bellemare,
  M.~G., Graves, A., Riedmiller, M., Fidjeland, A.~K., Ostrovski, G., et~al.
\newblock \href{https://www.nature.com/articles/nature14236}{Human-level
  control through deep reinforcement learning}.
\newblock \emph{Nature}, 518\penalty0 (7540):\penalty0 529--533, 2015.

\bibitem[Munos \& Szepesv{\'a}ri(2008)Munos and
  Szepesv{\'a}ri]{munos2008finite}
Munos, R. and Szepesv{\'a}ri, C.
\newblock \href{https://www.jmlr.org/papers/v9/munos08a.html}{Finite-Time
  Bounds for Fitted Value Iteration}.
\newblock \emph{Journal of Machine Learning Research}, 9\penalty0 (5):\penalty0
  815--857, 2008.

\bibitem[Munos et~al.(2016)Munos, Stepleton, Harutyunyan, and
  Bellemare]{munos2016safe}
Munos, R., Stepleton, T., Harutyunyan, A., and Bellemare, M.
\newblock
  \href{https://papers.nips.cc/paper/2016/hash/c3992e9a68c5ae12bd18488bc579b30d-Abstract.html}{Safe
  and efficient off-policy reinforcement learning}.
\newblock \emph{Advances in Neural Information Processing Systems}, 29, 2016.

\bibitem[Osband et~al.(2013)Osband, Russo, and Van~Roy]{osband2013more}
Osband, I., Russo, D., and Van~Roy, B.
\newblock
  \href{https://papers.nips.cc/paper/2013/hash/6a5889bb0190d0211a991f47bb19a777-Abstract.html}{(More)
  efficient reinforcement learning via posterior sampling}.
\newblock \emph{Advances in Neural Information Processing Systems}, 26, 2013.

\bibitem[Osband et~al.(2016{\natexlab{a}})Osband, Blundell, Pritzel, and
  Van~Roy]{osband2016deep}
Osband, I., Blundell, C., Pritzel, A., and Van~Roy, B.
\newblock
  \href{https://papers.nips.cc/paper/2016/hash/8d8818c8e140c64c743113f563cf750f-Abstract.html}{Deep
  exploration via bootstrapped DQN}.
\newblock \emph{Advances in Neural Information Processing Systems}, 29,
  2016{\natexlab{a}}.

\bibitem[Osband et~al.(2016{\natexlab{b}})Osband, Van~Roy, and
  Wen]{osband2016generalization}
Osband, I., Van~Roy, B., and Wen, Z.
\newblock \href{http://proceedings.mlr.press/v48/osband16.html}{Generalization
  and exploration via randomized value functions}.
\newblock In \emph{International Conference on Machine Learning}, pp.\
  2377--2386. PMLR, 2016{\natexlab{b}}.

\bibitem[Osband et~al.(2018)Osband, Aslanides, and
  Cassirer]{osband2018randomized}
Osband, I., Aslanides, J., and Cassirer, A.
\newblock
  \href{https://papers.nips.cc/paper/2018/hash/5a7b238ba0f6502e5d6be14424b20ded-Abstract.html}{Randomized
  prior functions for deep reinforcement learning}.
\newblock \emph{Advances in Neural Information Processing Systems}, 31, 2018.

\bibitem[Osband et~al.(2019)Osband, Van~Roy, Russo, and Wen]{osband2019deep}
Osband, I., Van~Roy, B., Russo, D.~J., and Wen, Z.
\newblock \href{https://jmlr.org/papers/v20/18-339.html}{Deep Exploration via
  Randomized Value Functions}.
\newblock \emph{Journal of Machine Learning Research}, 20\penalty0
  (124):\penalty0 1--62, 2019.

\bibitem[O’Donoghue et~al.(2018)O’Donoghue, Osband, Munos, and
  Mnih]{o2018uncertainty}
O’Donoghue, B., Osband, I., Munos, R., and Mnih, V.
\newblock \href{https://proceedings.mlr.press/v80/odonoghue18a.html}{The
  uncertainty bellman equation and exploration}.
\newblock In \emph{International Conference on Machine Learning}, pp.\
  3839--3848. PMLR, 2018.

\bibitem[Rashid et~al.(2020)Rashid, Peng, B{\"o}hmer, and
  Whiteson]{rashid2020optimistic}
Rashid, T., Peng, B., B{\"o}hmer, W., and Whiteson, S.
\newblock \href{https://openreview.net/forum?id=r1xGP6VYwH}{Optimistic
  Exploration even with a Pessimistic Initialisation}.
\newblock In \emph{International Conference on Learning Representations}, 2020.

\bibitem[Rasmussen(2005)]{Rasmussen2004}
Rasmussen, C.~E.
\newblock
  \emph{\href{https://direct.mit.edu/books/book/2320/Gaussian-Processes-for-Machine-Learning}{Gaussian
  processes for machine learning}}.
\newblock MIT press, 2005.

\bibitem[Riquelme et~al.(2018)Riquelme, Tucker, and Snoek]{riquelme2018deep}
Riquelme, C., Tucker, G., and Snoek, J.
\newblock \href{https://openreview.net/forum?id=SyYe6k-CW}{Deep Bayesian
  Bandits Showdown: An Empirical Comparison of Bayesian Deep Networks for
  Thompson Sampling}.
\newblock In \emph{International Conference on Learning Representations}, 2018.

\bibitem[Russo(2019)]{russo2019worst}
Russo, D.
\newblock
  \href{https://papers.nips.cc/paper/2019/hash/451ae86722d26a608c2e174b2b2773f1-Abstract.html}{Worst-case
  regret bounds for exploration via randomized value functions}.
\newblock \emph{Advances in Neural Information Processing Systems}, 32, 2019.

\bibitem[Savinov et~al.(2018)Savinov, Raichuk, Vincent, Marinier, Pollefeys,
  Lillicrap, and Gelly]{savinov2018episodic}
Savinov, N., Raichuk, A., Vincent, D., Marinier, R., Pollefeys, M., Lillicrap,
  T., and Gelly, S.
\newblock \href{https://openreview.net/forum?id=SkeK3s0qKQ}{Episodic Curiosity
  through Reachability}.
\newblock In \emph{International Conference on Learning Representations}, 2018.

\bibitem[Schaul et~al.(2016)Schaul, Quan, Antonoglou, and Silver]{schaul2016}
Schaul, T., Quan, J., Antonoglou, I., and Silver, D.
\newblock \href{http://arxiv.org/abs/1511.05952}{Prioritized Experience
  Replay}.
\newblock In \emph{International Conference on Learning Representations}, 2016.

\bibitem[Schrittwieser et~al.(2020)Schrittwieser, Antonoglou, Hubert, Simonyan,
  Sifre, Schmitt, Guez, Lockhart, Hassabis, Graepel,
  et~al.]{Schrittwieser2019-df}
Schrittwieser, J., Antonoglou, I., Hubert, T., Simonyan, K., Sifre, L.,
  Schmitt, S., Guez, A., Lockhart, E., Hassabis, D., Graepel, T., et~al.
\newblock \href{https://www.nature.com/articles/s41586-020-03051-4}{Mastering
  atari, go, chess and shogi by planning with a learned model}.
\newblock \emph{Nature}, 588\penalty0 (7839):\penalty0 604--609, 2020.

\bibitem[Snoek et~al.(2015)Snoek, Rippel, Swersky, Kiros, Satish, Sundaram,
  Patwary, Prabhat, and Adams]{snoek2015scalable}
Snoek, J., Rippel, O., Swersky, K., Kiros, R., Satish, N., Sundaram, N.,
  Patwary, M., Prabhat, M., and Adams, R.
\newblock \href{http://proceedings.mlr.press/v37/snoek15.html}{Scalable
  bayesian optimization using deep neural networks}.
\newblock In \emph{International Conference on Machine Learning}, pp.\
  2171--2180. PMLR, 2015.

\bibitem[Tiapkin et~al.(2022)Tiapkin, Belomestny, Moulines, Naumov, Samsonov,
  Tang, Valko, and Menard]{tiapkin2022dirichlet}
Tiapkin, D., Belomestny, D., Moulines, E., Naumov, A., Samsonov, S., Tang, Y.,
  Valko, M., and Menard, P.
\newblock From dirichlet to rubin: Optimistic exploration in rl without
  bonuses.
\newblock In \emph{International Conference on Machine Learning}. PMLR, 2022.

\bibitem[Toromanoff et~al.(2019)Toromanoff, Wirbel, and Moutarde]{recordmean}
Toromanoff, M., Wirbel, E., and Moutarde, F.
\newblock \href{https://arxiv.org/abs/1908.04683}{Is Deep Reinforcement
  Learning Really Superhuman on Atari? Leveling the playing field}.
\newblock \emph{arXiv preprint arXiv:1908.04683}, 2019.

\bibitem[Tziortziotis et~al.(2013)Tziortziotis, Dimitrakakis, and
  Blekas]{tziortziotis2013linear}
Tziortziotis, N., Dimitrakakis, C., and Blekas, K.
\newblock \href{https://www.ijcai.org/Proceedings/13/Papers/255.pdf}{Linear
  Bayesian reinforcement learning}.
\newblock In \emph{International joint conference on artificial intelligence},
  2013.

\bibitem[Vinyals et~al.(2019)Vinyals, Babuschkin, Czarnecki, Mathieu, Dudzik,
  Chung, Choi, Powell, Ewalds, Georgiev, et~al.]{vinyals2019grandmaster}
Vinyals, O., Babuschkin, I., Czarnecki, W.~M., Mathieu, M., Dudzik, A., Chung,
  J., Choi, D.~H., Powell, R., Ewalds, T., Georgiev, P., et~al.
\newblock \href{https://www.nature.com/articles/s41586-019-1724-z}{Grandmaster
  level in StarCraft II using multi-agent reinforcement learning}.
\newblock \emph{Nature}, 575\penalty0 (7782):\penalty0 350--354, 2019.

\bibitem[Zanette \& Brunskill(2019)Zanette and Brunskill]{zanette2019tighter}
Zanette, A. and Brunskill, E.
\newblock \href{https://proceedings.mlr.press/v97/zanette19a.html}{Tighter
  problem-dependent regret bounds in reinforcement learning without domain
  knowledge using value function bounds}.
\newblock In \emph{International Conference on Machine Learning}, pp.\
  7304--7312. PMLR, 2019.

\end{thebibliography}
\bibliographystyle{icml2023}


\appendix
\onecolumn

\section{Atari benchmark: average returns} \label{app:numericalscores}

\begin{table}[h!]
\caption{
Average return for evaluation episodes in each individual game (1M environment steps, averaged across three seeds).
}
\label{table:scores}
\vskip 0.15in
\centering
{
\footnotesize
\begin{tabular}{lcccc}

\toprule
\textbf{Game} &  \textbf{PSDRL} & \textbf{B+P} & \textbf{SU} & \textbf{Dv2} \\
\midrule    
Alien & \textbf{1199} & 505& 386 & 844 \\
Amidar & \textbf{156} & 36& 19 & 117 \\
Assault & 662 & 461& 147 & \textbf{1589} \\
Asterix & \textbf{1235} & 514& 431 & 1090 \\
Asteroids & \textbf{1141} & 932& 760 & 549 \\
Atlantis & 35273 & 20578& 11174 & \textbf{163070} \\
Bank Heist & 596 & 125& 18 & \textbf{673} \\
Battle Zone & \textbf{10612} & 8127& 6007 & 3275 \\
Beam Rider & 779 & \textbf{2091}& 370 & 587 \\
Berzerk & 386 & 226& \textbf{33644} & 339 \\
Bowling & 15 & 5& 4 & \textbf{36} \\
Boxing & \textbf{79} & 6& -46 & 73 \\
Breakout & \textbf{46} & 8& 0 & 2 \\
Centipede & 2594 & 2446& 2888 & \textbf{4067} \\
Chopper Command & 899 & 502& 589 & \textbf{1289} \\
Crazy Climber & 30392 & 17942& 2086 & \textbf{65140} \\
Demon Attack & 410 & 1225& 776 & \textbf{1233} \\
Double Dunk & -16 & -6& \textbf{-1} & -4 \\
Enduro & \textbf{363} & 0& 0 & 178 \\
Fishing Derby & \textbf{-60} & -86& -96 & -85 \\
Freeway & \textbf{28} & 1& 7 & \textbf{28} \\
Frostbite & \textbf{929} & 204& 6 & 354 \\
Gopher & 2683 & 865& 316 & \textbf{5194} \\
Gravitar & \textbf{494} & 92& 208 & 222 \\
Hero & \textbf{7965} & 2202& 768 & 3096 \\
Ice Hockey & -12 & -13& -13 & \textbf{-6} \\
James bond & 231 & 69& 41 & \textbf{470} \\
Kangaroo & \textbf{3180} & 384& 190 & 3063 \\
Krull & \textbf{10802} & 2235& 903 & 7301 \\
Kung Fu Master & 17528 & 6647& 412 & \textbf{22249} \\
Montezuma Revenge & 0 & 0& \textbf{7} & 0 \\
Ms Pacman & \textbf{1824} & 812& 429 & 1502 \\
Name This Game & 4037 & 3767& 2473 & \textbf{6534} \\
Phoenix & \textbf{3876} & 1756& 323 & 1981 \\
Pitfall & -44 & -63& -469 & \textbf{-39} \\
Pong & \textbf{11} & -20& -21 & 0 \\
Private Eye & 68 & -53& -793 & \textbf{79} \\
Qbert & \textbf{4245} & 560& 337 & 2188 \\
Riverraid & 3858 & 2488& 1235 & \textbf{4174} \\
Road Runner & \textbf{22272} & 528& 984 & 11954 \\
Robotank & 3 & 5& 2 & \textbf{10} \\
Seaquest & 1070 & 307& 96 & \textbf{1282} \\
Skiing & \textbf{-14980} & -28748& -22105 & -19777 \\
Solaris & 1794 & 396& \textbf{2083} & 866 \\
Space Invaders & \textbf{511} & 240& 335 & 403 \\
Star Gunner & \textbf{2542} & 1349& 531 & 1188 \\
Tennis & -22 & -17& -13 & \textbf{-10} \\
Time Pilot & \textbf{4156} & 2996& 2390 & 2767 \\
Tutankham & \textbf{94} & 60& 19 & 89 \\
Up N Down & 8596 & 1587& 794 & \textbf{90684} \\
Venture & 0 & 3& 0 & \textbf{8} \\
Video Pinball & 8857 & 1145& 8641 & \textbf{14973} \\
Wizard Of Wor & 1459 & 848& 1181 & \textbf{2124} \\
Yars Revenge & \textbf{17038} & 3657& 1804 & 12144 \\
Zaxxon & \textbf{4413} & 1158& 206 & 2671 \\
\bottomrule
 \end{tabular}
}
\vskip -0.1in
\end{table}  

\section{Atari benchmark: learning curves}\label{app:learningcurves}

\begin{figure}[H]
\vskip 0.2in
\begin{center}
\centerline{\includegraphics[width=0.85\linewidth]{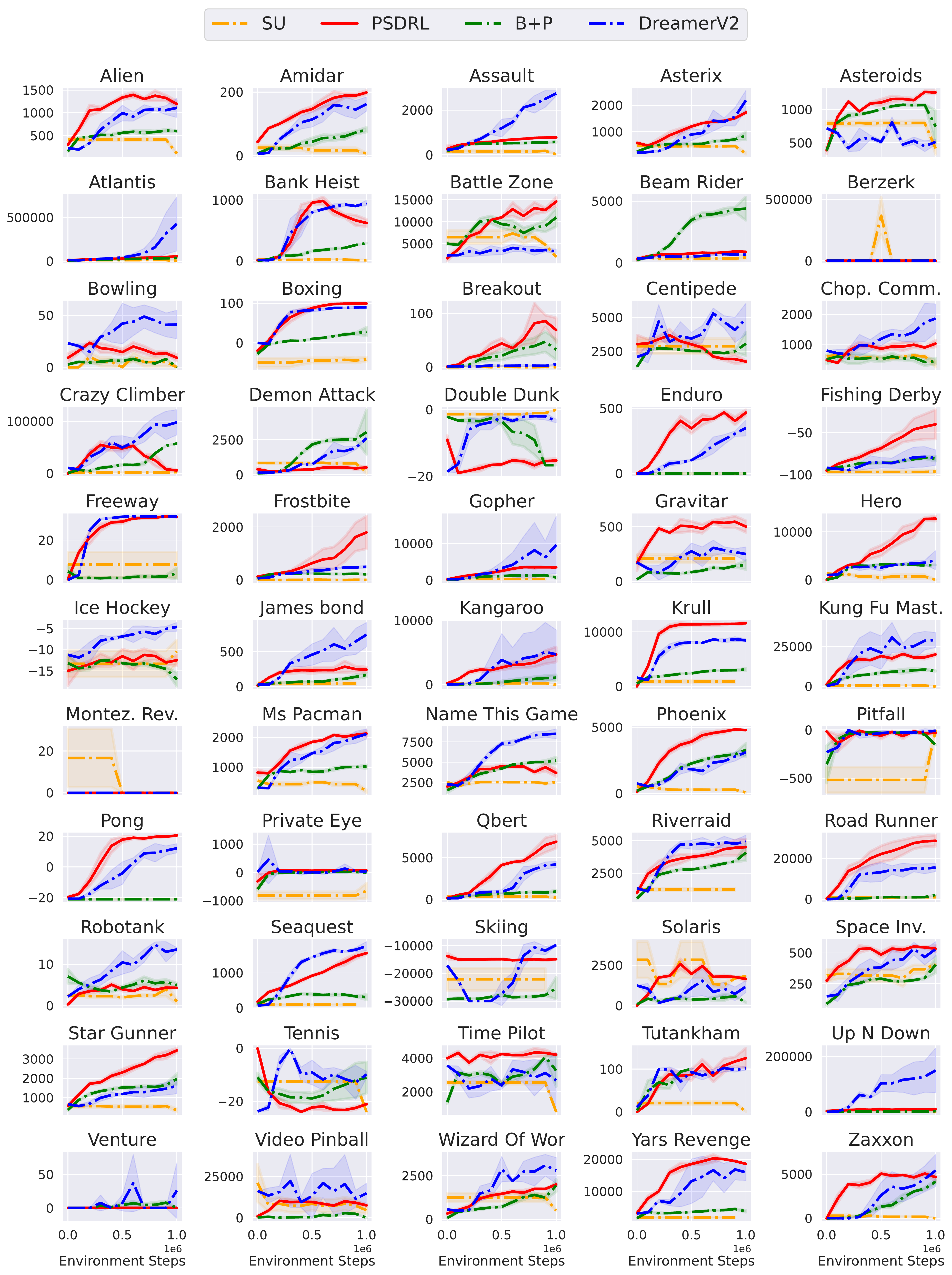}}
\caption{Returns for evaluation episodes in each individual game (1M environment steps, averaged across three seeds).
}
\label{all_games_curves_raw}
\end{center}
\vskip -0.1in
\end{figure}

\newpage
\section{Normalized score visualizations}
\label{sec:normalized_visuals}

\begin{figure*}[h]
\vskip 0.2in
\begin{center}
    \includegraphics[width=0.95\linewidth]{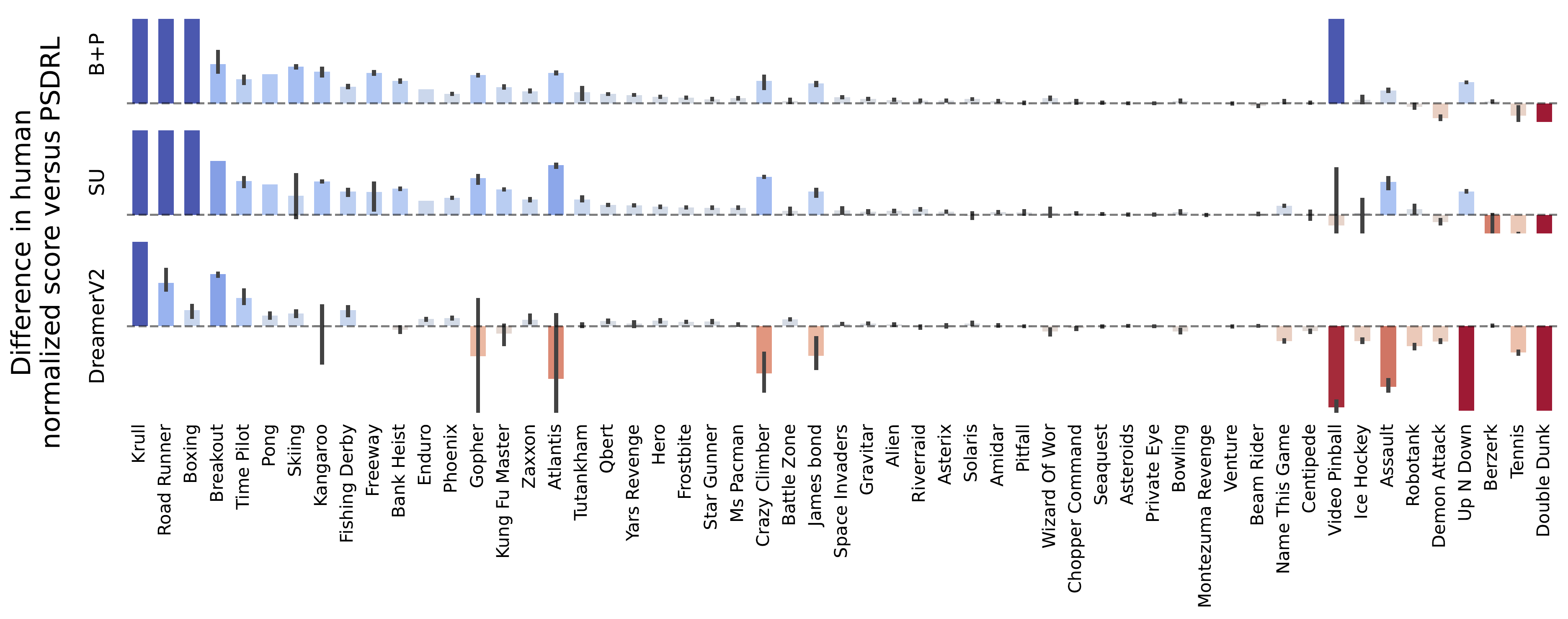}
    \caption{A comparison between PSDRL and B+P (top), SU (middle), and Dv2 (bottom) in terms of human-normalized score. Blue indicates that PSDRL scores higher than the baseline, red the opposite. Y-axis has been clipped to $[-2.5, 2.5]$ to facilitate visualization. This figure is identical to Figure \ref{fig:all_scores} and is presented here to facilitate comparison with Figure \ref{fig:clipped}.}
    \label{fig:recprd}
\end{center}
\vskip -0.2in
\end{figure*}

\begin{figure*}[h]
\vskip 0.2in
\begin{center}
    \includegraphics[width=0.95\linewidth]{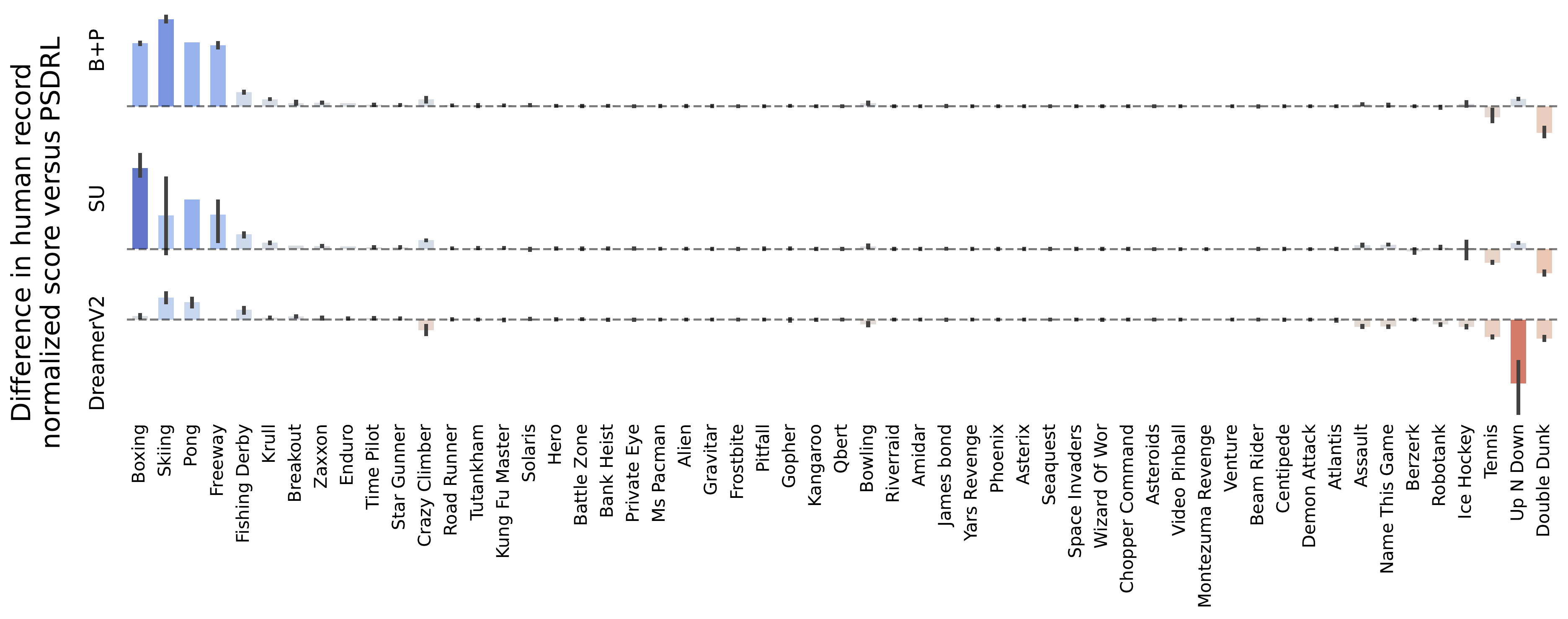}
    \caption{A comparison between PSDRL and B+P (top), SU (middle), and Dv2 (bottom) in terms of record-normalized score. Blue indicates that PSDRL scores higher than the baseline, red the opposite.}
    \label{fig:clipped}
\end{center}
\vskip -0.1in
\end{figure*}

Figures \ref{fig:recprd} and \ref{fig:clipped} compare PSDRL with the baselines on 55 Atari games in terms of scores normalized with respect to a representative human performance or the human record performance, respectively.
Concretely, the height of each bar corresponds to the difference between the normalized score of PSDRL and the normalized score of a given baseline. The vertical black lines represent $95\%$ bootstrapped confidence intervals.
Darker shades of blue correspond to a wider gap in favor of PSDRL, darker shades of red correspond to a wider gap in favor of a given baseline. In Figure \ref{fig:recprd}, the height of each bar has been clipped to the interval $[-2.5, 2.5]$, since otherwise most bars would be difficult to see.


\newpage
\section{Hyperparameters and implementation details}\label{app:hyperparams}

Hyperparameters search was conducted through a grid search over a carefully selected subset of hyperparameters for each algorithm.
The evaluation metric employed for this search (average return during evaluation episodes) is further averaged over results on six Atari games that are commonly used for this purpose \citep{munos2016safe, janz2019successor}: \textsc{Asterix}, \textsc{Enduro}, \textsc{Freeway}, \textsc{Hero}, \textsc{Qbert}, and \textsc{Seaquest}.
In order to guarantee a fair comparison, we chose hyperparameters that allow each algorithm to finish training for each game in under 10 hours.


\subsection{Posterior sampling for deep reinforcement learning}

\begin{figure*}[h]
\vskip 0.2in
\begin{center}
    \includegraphics[width=0.45\linewidth]{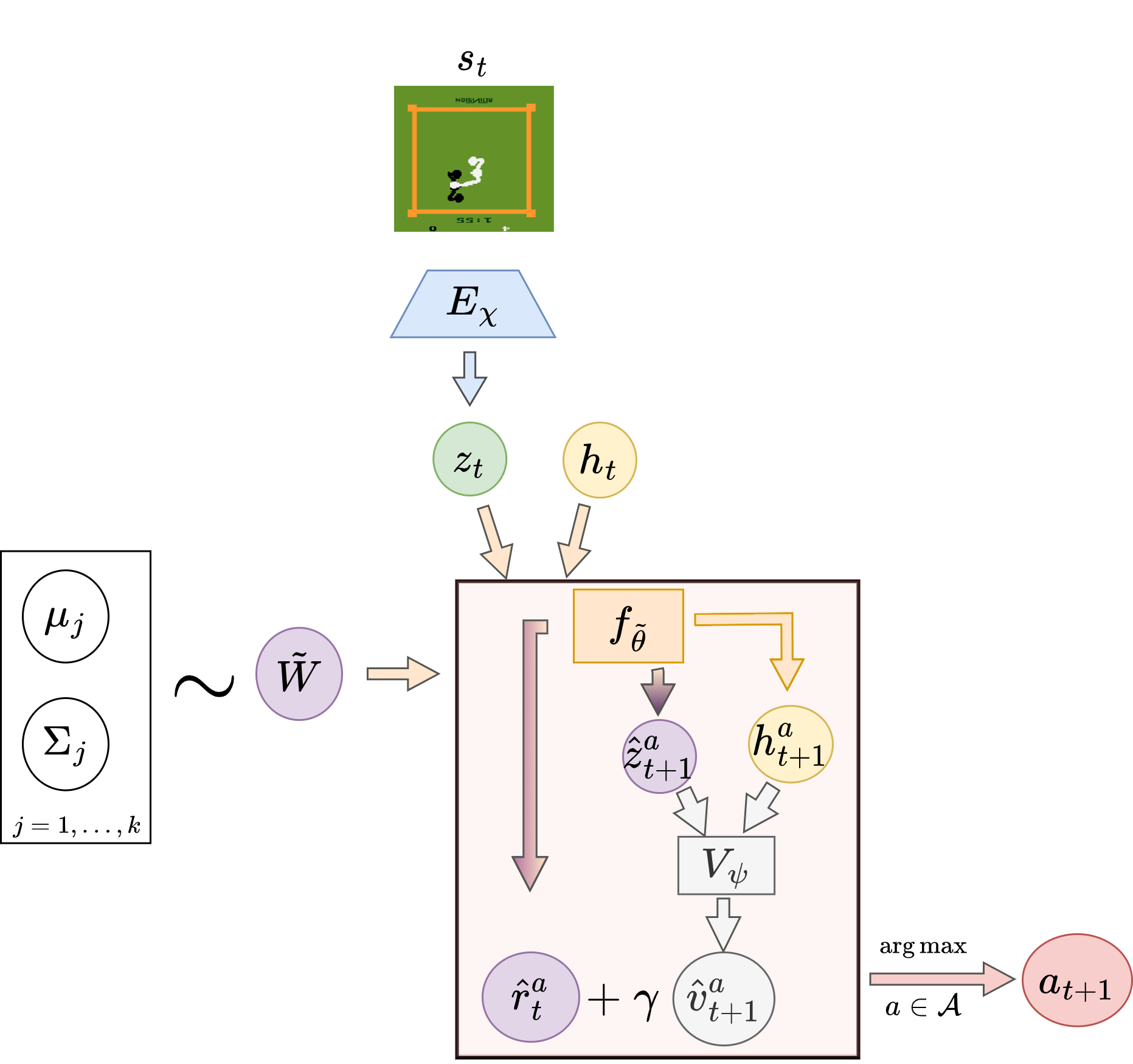}
    \caption{Diagram of the interactions between components in PSDRL when interacting with the environment. At timestep  $t$, the agent encodes an observation $s_t$ to latent state $z_t$. Given $z_t$ and the previous hidden state $h_t$, the forward model $f_{\tilde{\theta}}$ with sampled parameters $\tilde{W}$ predicts the next latent state $\hat{z}^a_{t+1}$ and reward $\hat{r}^a_t$ for every action $a \in \mathcal{A}$. For each of the predicted next latent states, the value function $V_\psi$ predicts a valuation $\hat{v}^a_{t+1}$. Finally, an action is yielded by computing: $a_{t+1} =  \argmax_{a} \left[ \hat{r}^{(a)}_{t} + \gamma \hat{v}_{t+1}^{(a)} \right]$, and the corresponding hidden state $h_{t+1}^a$ is carried over to the next timestep.}
    \label{fig:int_diag}
\end{center}
\vskip -0.2in
\end{figure*}

The implementation for PSDRL can be found at \url{https://github.com/remosasso/PSDRL}. Table \ref{tab:psrl-hparams} presents the hyperparameters for PSDRL, the search sets used for grid search, and the resulting values used for the experiments. Additionally, note that we update the components more frequently ($m=250$) during the first 100k environment steps in comparison with the remaining steps ($m=1000$). See Figure \ref{fig:int_diag} for a diagram that demonstrates how the components of PSDRL interact when interacting with the environment.

PSDRL receives states represented as $64 \times 64$ grayscale images and does not require frame stacking due to employing a recurrent forward model. The autoencoder model, forward model, termination model, and value network are trained with the Adam optimizer \citep{kingma2014adam} using a learning rate of 1e-4.

In our implementation, a batch $\mathcal{B}$ containing $B \times L$ transitions is sampled and used in slightly different ways by each component in a given training iteration. The autoencoder model performs $B$ gradient updates with an \emph{inner batch} of size $L$. The forward model, termination model, and value network perform $L/l$ gradient updates with a batch of size $B$, where $l$ denotes the horizon for truncated backpropagation through time. For the forward and termination models, $l=4$. For the value network, $l=1$. Therefore, according to the parameters in Table \ref{tab:psrl-hparams}, the parameters of the value network are updated $\kappa L/l = 750$ times with batches of size $B = 125$ during each planning iteration (Alg. \ref{PSDRL}, line 12).

Instead of choosing actions strictly according to the greedy policy (Eq. \ref{eq:greedypolicy}), our implementation of PSDRL chooses actions according to the corresponding $\hat{\epsilon}$-greedy policy, where $\hat{\epsilon}$ is a very small value.
This ensures that the agent does not waste all of its time caught in a loop between two states early in training even if the sampled model would justify this behavior.

We make use of an NVIDIA A100 GPU for training. PSDRL takes about 9 hours per 1 million environment steps per game.

\begin{table}[h]
\caption{Hyperparameters for PSDRL.}
\label{tab:psrl-hparams}
\vskip 0.15in
\begin{center}
\small
\begin{tabular}{lccc}
\toprule
\textbf{Name} & \textbf{Symbol} & \textbf{Search set} & \textbf{Value used} \\
\midrule
\textbf{Bayesian linear regression} \\
\hspace{2mm} Prior variance for latent state parameters & $\sigma^2_S$ & \{1e1, 1e3, 1e5\} & 1e3 \\
\hspace{2mm} Prior variance for reward parameters & $\sigma^2_R$ & \{1e1, 1e3, 1e5\} & 1e3 \\
\hspace{2mm} Noise variance & $\sigma^2$ & --- & 1\\
\midrule
\textbf{Forward model} \\
\hspace{2mm} Number of layers & & --- & 5 \\
\hspace{2mm} Activation function & & --- & Tanh \\
\hspace{2mm} Hidden units & & --- & 2292 \\
\hspace{2mm} Learning rate & &--- & 1e-4 \\
\hspace{2mm} Training iterations & &--- & 3\\
\hspace{2mm} Recurrent hidden units & &--- & 756\\
\hspace{2mm} Window update length && $l$ &4\\
\midrule
\textbf{Terminal model}\\
\hspace{2mm} Number of layers & & --- & 4 \\
\hspace{2mm} Activation function & & --- & Tanh \\
\hspace{2mm} Hidden units & & --- & 1536 \\
\hspace{2mm} Learning rate & &--- & 1e-4 \\
\hspace{2mm} Training iterations & &--- & 3\\
\midrule
\textbf{Value network}\\
\hspace{2mm} Number of layers & & --- & 5 \\
\hspace{2mm} Activation function & & --- & Tanh \\
\hspace{2mm} Hidden units & & --- & 2292 \\
\hspace{2mm} Learning rate & &--- & 1e-4 \\
\hspace{2mm} Training iterations & $\kappa$ &--- & 3\\
\hspace{2mm} Target update frequency &  & 4\\
\hspace{2mm} Discount factor & $\gamma$ & \{0.99, 0.999\} & 0.99 \\
\midrule
\textbf{Autoencoder model}\\
\hspace{2mm} Number of encoder layers & & --- & 4 \\
\hspace{2mm} Number of decoder layers & & --- & 4 \\
\hspace{2mm} Activation function & & --- & ReLu \\
\hspace{2mm} Encoded dimensions & & --- & 1536 \\
\hspace{2mm} Learning rate & &--- & 1e-4 \\
\hspace{2mm} Training iterations & & --- &3\\
\midrule
\textbf{Replay buffer}\\
\hspace{2mm} Batch size & $B$ & \{125, 250\} & 125\\
\hspace{2mm} Sequence length & $L$ & ---   &250\\
\hspace{2mm} Capacity & $C$ &--- & 1e5 \\
\midrule
\textbf{Environment interaction}\\
\hspace{2mm} Update frequency & $m$ & ---  & $1000$\\
\hspace{2mm} Policy noise & $\hat{\epsilon}$ & 0.001 \\
\bottomrule
\end{tabular}
\end{center}
\vskip -0.1in
\end{table}

\clearpage

\subsection{Successor uncertainties}

Via private communication, the authors of SU suggested that three hyperparameters should be tuned to improve the sample efficiency of their algorithm. These hyperparameters are reported in Table \ref{tab:hparamsSU}.

We make use of the implementation published by the authors at \url{https://github.com/DavidJanz/successor_uncertainties_atari}.

\begin{table}[H]
\caption{Hyperparameters for SU.}
\vskip 0.15in
\begin{center}
\small
\begin{tabular}{lcc}
\toprule
\textbf{Name} & \textbf{Search set} & \textbf{Value used} \\
\midrule
Likelihood prior & \{1e-1, 1e-3, 1e-5\} & 1e-3 \\
Training interval & \{2, 4\} & 2 \\ 
Batch size & \{32, 64\} & 64 \\
\bottomrule
\end{tabular}
\end{center}
\vskip -0.1in
\label{tab:hparamsSU}
\end{table}

\subsection{Bootstrapped DQN with randomized priors}

Because B+P shares several components with SU, we selected a similar set of hyperparameters to tune for sample efficiency. These hyperparameters are reported in Table \ref{tab:hparamsBDQN}.

We make use of an implementation for Atari available at \url{https://github.com/johannah/bootstrap_dqn}.

\begin{table}[H]
\caption{Hyperparameters for B+Q.}
\vskip 0.15in
\begin{center}
\small
\begin{tabular}{lcc}
\toprule
\textbf{Name} & \textbf{Search set} & \textbf{Value used} \\
\midrule
Likelihood prior & \{1,3,10\} & 1 \\
Training interval & \{2, 4\} & 2 \\ 
Batch size & \{32, 64\} & 64 \\
\bottomrule
\end{tabular}
\end{center}
\vskip -0.1in
\label{tab:hparamsBDQN}
\end{table}

\subsection{DreamerV2} 
\label{app:dv2}
The authors of Dv2 mention that the training interval (originally 16) should be decreased \citep{Hafner2020-ry} to increase sample efficiency. Table \ref{tab:hparamsDreamer} reports the corresponding values employed for grid search.

We make use of the implementation published by the authors at \url{https://github.com/danijar/dreamerv2}.

\begin{table}[H]
\centering
\caption{Hyperparameters for Dv2.}
\vskip 0.15in
\small
\begin{tabular}{lcc}
\toprule
\textbf{Name} & \textbf{Search set} & \textbf{Value used} \\
\midrule
Training interval  & \{4,8,12\} & 8 \\
\bottomrule
\end{tabular}
\label{tab:hparamsDreamer}
\end{table}
\newpage

\section{Runtime comparison}
Table \ref{tab:wall-clock} reports the amount of wall clock time required for 1M environment steps in 7 games for each of the algorithm implementations used in this paper. Note that this is the computational efficiency on an NVIDIA A100 GPU \emph{after} tuning the baselines for data efficiency.

\begin{table}[h]
\caption{Wall clock time for 1M environment steps for each of the algorithms after tuning.}
\centering
\label{tab:wall-clock}
\vskip 0.15in
\begin{tabular}{lllll}
\toprule
\textbf{Game} & \hphantom{hello} \textbf{SU}& \hphantom{heo} \textbf{PSDRL} & \hphantom{hello}  \textbf{B+P}  &  \hphantom{hello} \textbf{Dv2}\\
\midrule
Freeway & 5h3m $\pm$ 20m & 8h53m $\pm$ 0m & 8hr11m $\pm$ 24m & 11h20m $\pm$ 31m \\
Qbert & 6h28m $\pm$ 23m & 7h39m $\pm$ 43m & 9h43m $\pm$ 9m & 10h52m $\pm$ 20m \\ 
Enduro & 5h13m $\pm$ 25m & 9h35m $\pm$ 15m & 8h12m $\pm$ 32m & 10h58m $\pm$ 3m \\ 
Asterix & 4h54m $\pm$ 4m & 7h44m $\pm$ 24m & 9h7m $\pm$ 14m & 10h18m $\pm$ 18m \\ 
Seaquest & 5h30m $\pm$ 27 m & 8h31m $\pm$ 2m & 8h56m $\pm$ 15m & 10h33m $\pm$ 7m \\ 
Pong & 7h0m $\pm$ 34m & 7h58m $\pm$ 42m & 8h58m $\pm$ 6m & 11h7m $\pm$ 35m \\ 
Hero & 5h46m $\pm$ 19m & 9h26m $\pm$ 2m & 9h32m $\pm$ 28m & 12h24m $\pm$ 26m \\ 
\textbf{Average} & \textbf{5h41m} $\pm$ \textbf{64m} & \textbf{8h31m} $\pm$ \textbf{44m} & \textbf{8h56m} $\pm$ \textbf{32m} & \textbf{11h3m} $\pm$ \textbf{37m}\\
\bottomrule
\end{tabular}
\end{table}


\clearpage

\section{Decoded sampled forward model predictions} \label{app:reconstructions}
\begin{figure*}[h]
    \centering
    \vskip 0.2in
    \begin{subfigure}[t]{\linewidth}
        \centering
        \vskip 0.05in
        \rotatebox{90}{\hphantom{hll}\small True} \hskip 0.05in
        \includegraphics[width=0.8\linewidth]{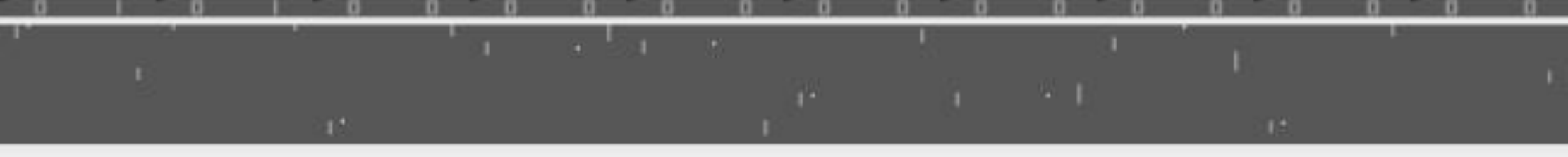}\\
        \vskip 0.05in
        \rotatebox{90}{\hphantom{hl}\small Model} \hskip 0.05in   
        \includegraphics[width=0.8\linewidth]{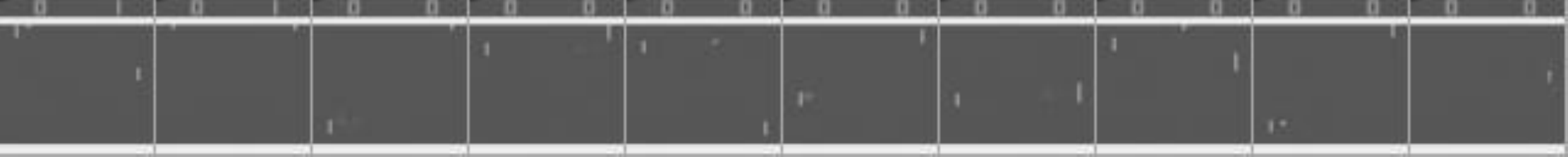}\\
        \vskip 0.05in
        \rotatebox{90}{\hphantom{hll}\small Error}   \hskip 0.05in
        \includegraphics[width=0.8\linewidth]{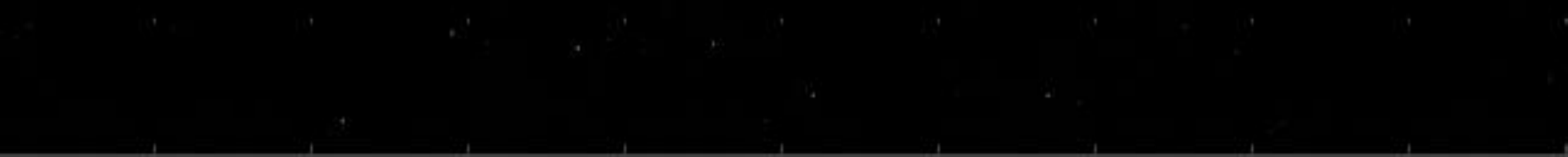}
    \end{subfigure}
    \vskip 0.1in
    \begin{subfigure}[t]{\linewidth}
        \centering
        \vskip 0.05in
        \rotatebox{90}{\hphantom{hll}\small True}   \hskip 0.05in     
        \includegraphics[width=0.8\linewidth]{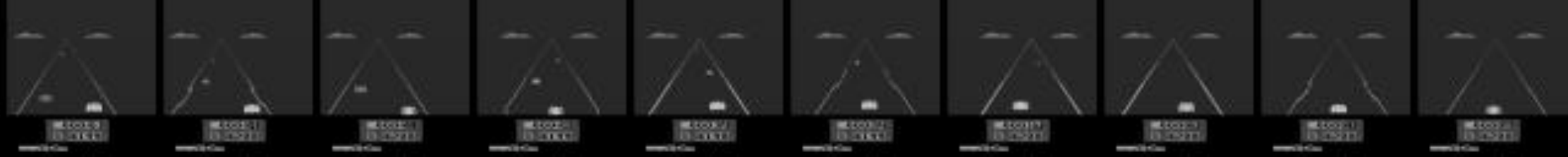}\\
        \vskip 0.05in
        \rotatebox{90}{\hphantom{hl}\small Model} \hskip 0.05in   
        \includegraphics[width=0.8\linewidth]{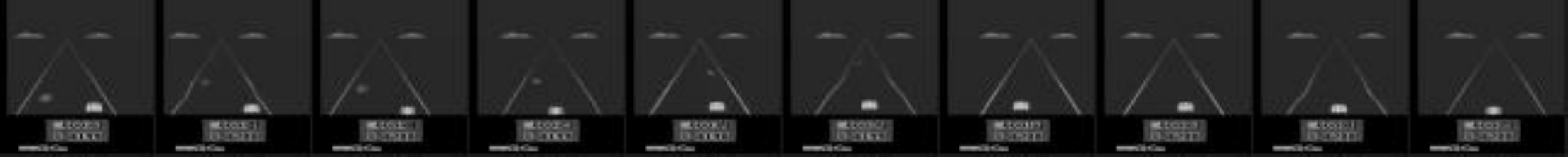}\\
        \vskip 0.05in
        \rotatebox{90}{\hphantom{hll}\small Error}   \hskip 0.05in
        \includegraphics[width=0.8\linewidth]{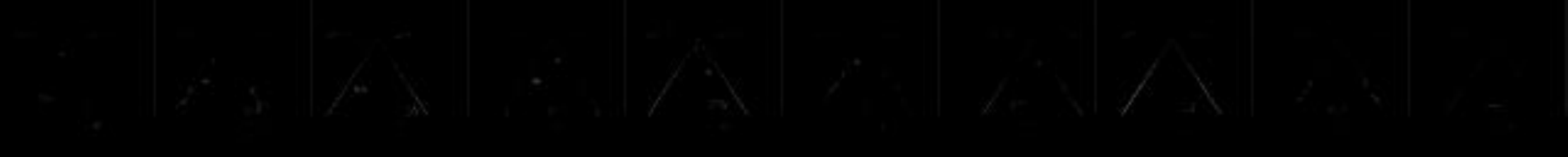}
    \end{subfigure}
    \vskip 0.1in
    \begin{subfigure}[t]{\linewidth}
        \centering
        \vskip 0.05in
        \rotatebox{90}{\hphantom{hll}\small True}    \hskip 0.05in    
        \includegraphics[width=0.8\linewidth]{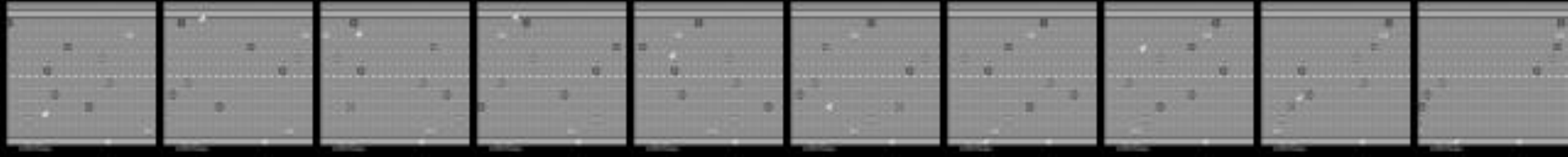}\\
        \vskip 0.05in
        \rotatebox{90}{\hphantom{hl}\small Model} \hskip 0.05in   
        \includegraphics[width=0.8\linewidth]{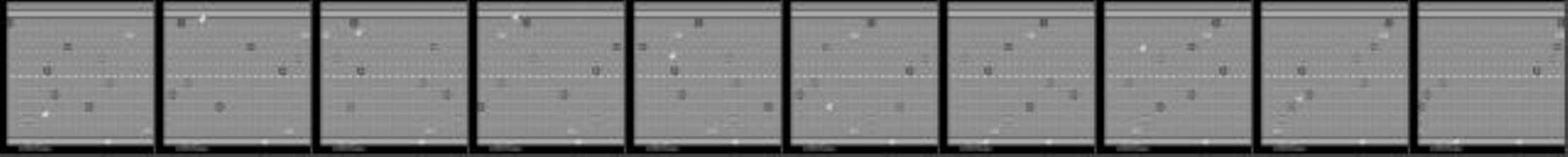}\\
        \vskip 0.05in
        \rotatebox{90}{\hphantom{hll}\small Error}   \hskip 0.05in
        \includegraphics[width=0.8\linewidth]{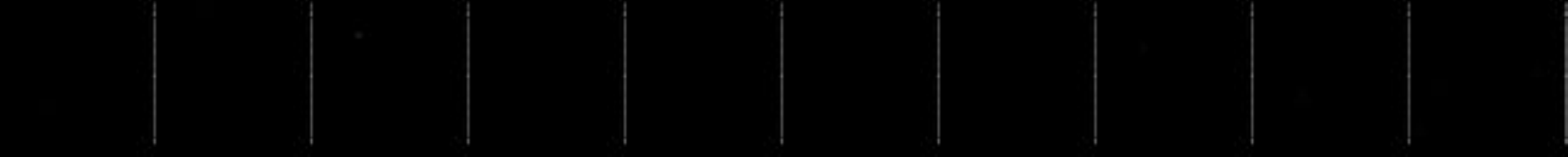}
    \end{subfigure}
    \vskip 0.1in

    \caption{Decoded sampled forward model predictions compared to the true environment state for \textsc{Pong, Enduro,} and \textsc{Freeway}. }
    \vskip -0.1in
    \label{fig:reconstrpart1}
\end{figure*}

\begin{figure*}[h]
    \centering
    \vskip 0.2in

    \begin{subfigure}[t]{\linewidth}
        \centering
        \vskip 0.05in
        \rotatebox{90}{\hphantom{hll}\small True}      \hskip 0.05in  
        \includegraphics[width=0.8\linewidth]{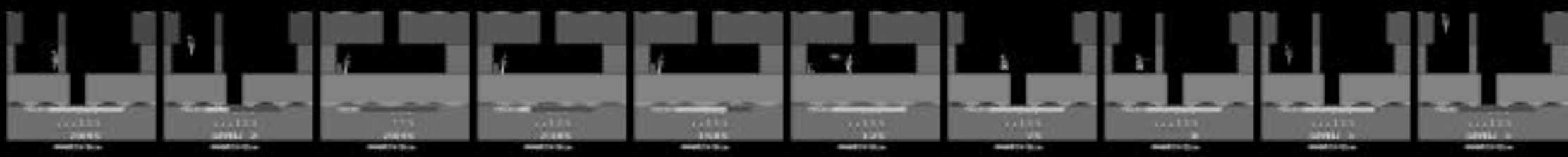}\\
        \vskip 0.05in
        \rotatebox{90}{\hphantom{hl}\small Model}   \hskip 0.05in
        \includegraphics[width=0.8\linewidth]{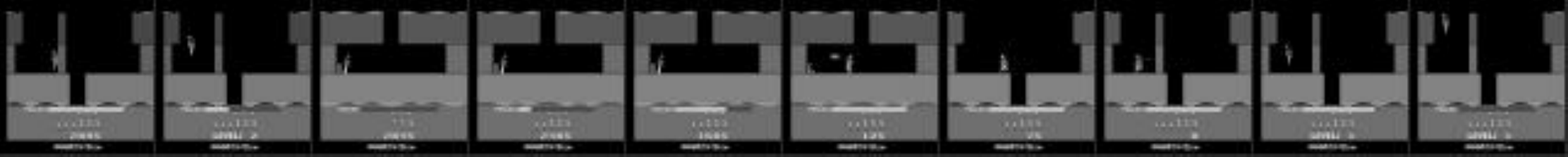}\\
        \vskip 0.05in
        \rotatebox{90}{\hphantom{hll}\small Error}   \hskip 0.05in
        \includegraphics[width=0.8\linewidth]{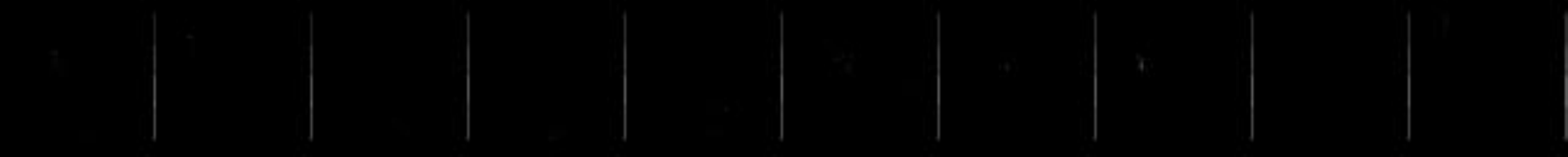}
    \end{subfigure}
    \vskip 0.1in
    \begin{subfigure}[t]{\linewidth}
        \centering
        \vskip 0.05in
        \rotatebox{90}{\hphantom{hll}\small True}   \hskip 0.05in     
        \includegraphics[width=0.8\linewidth]{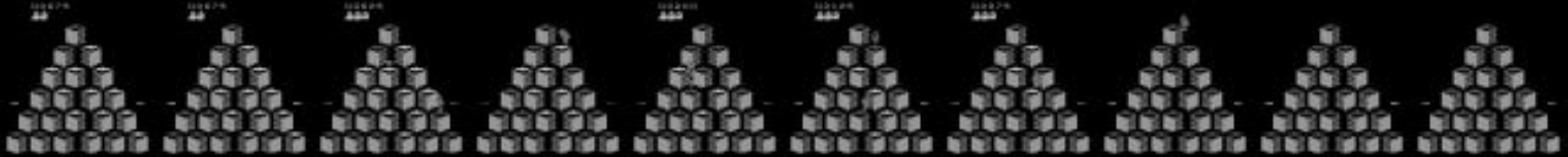}\\
        \vskip 0.05in
        \rotatebox{90}{\hphantom{hl}\small Model} \hskip 0.05in   
        \includegraphics[width=0.8\linewidth]{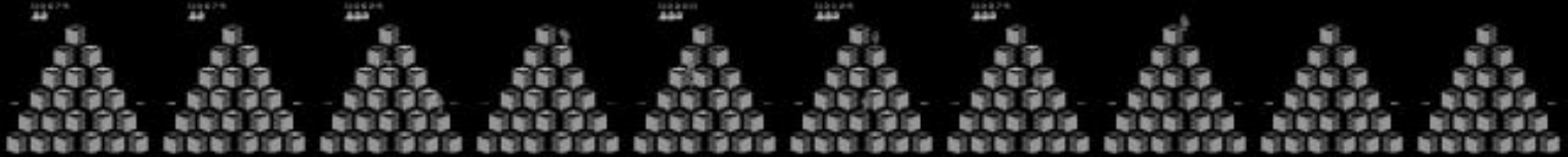}\\
        \vskip 0.05in
        \rotatebox{90}{\hphantom{hll}\small Error}   \hskip 0.05in
        \includegraphics[width=0.8\linewidth]{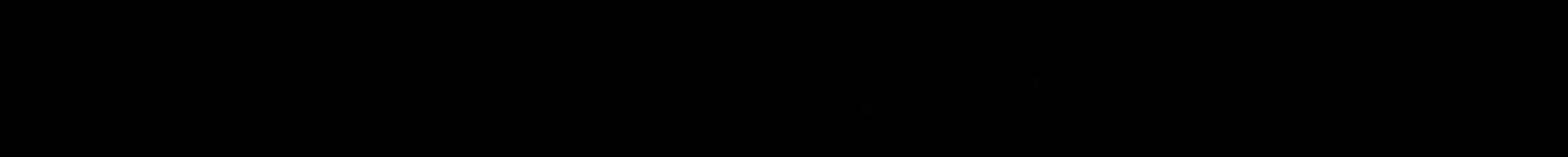}
    \end{subfigure}
    \vskip 0.1in
    \begin{subfigure}[t]{\linewidth}
        \centering
        \vskip 0.05in
        \rotatebox{90}{\hphantom{hll}\small True}    \hskip 0.05in    
        \includegraphics[width=0.8\linewidth]{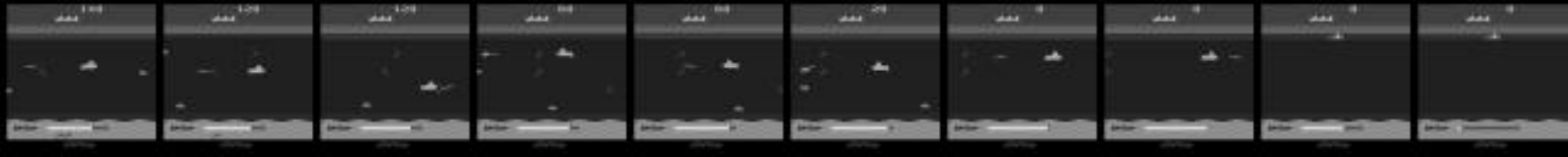}\\
        \vskip 0.05in
        \rotatebox{90}{\hphantom{hl}\small Model} \hskip 0.05in   
        \includegraphics[width=0.8\linewidth]{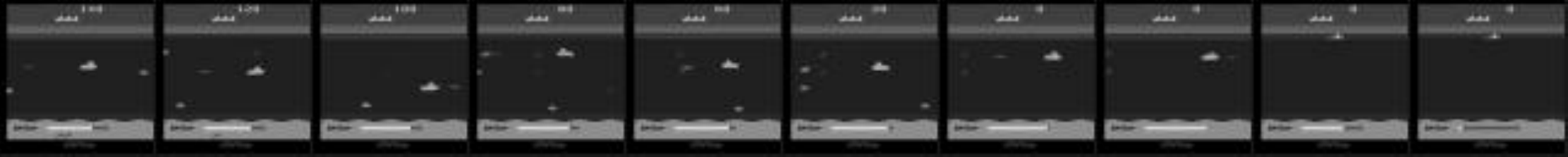}\\
        \vskip 0.05in
        \rotatebox{90}{\hphantom{hll}\small Error}   \hskip 0.05in
        \includegraphics[width=0.8\linewidth]{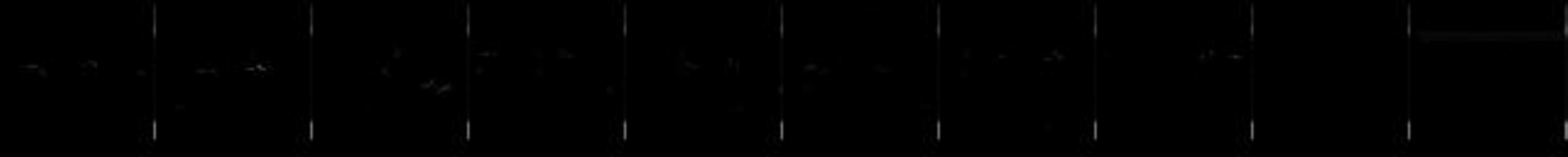}
    \end{subfigure}
    \vskip 0.1in
    \begin{subfigure}[t]{\linewidth}
        \centering
        \vskip 0.05in
        \rotatebox{90}{\hphantom{hll}\small True}    \hskip 0.05in    
        \includegraphics[width=0.8\linewidth]{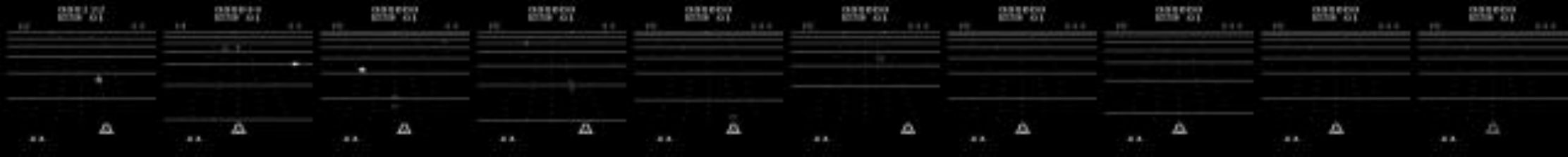}\\
        \vskip 0.05in
        \rotatebox{90}{\hphantom{hl}\small Model} \hskip 0.05in   
        \includegraphics[width=0.8\linewidth]{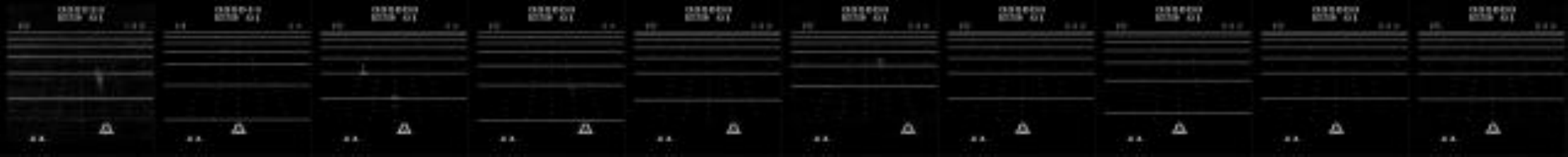}\\
        \vskip 0.05in
        \rotatebox{90}{\hphantom{hll}\small Error}   \hskip 0.05in
        \includegraphics[width=0.8\linewidth]{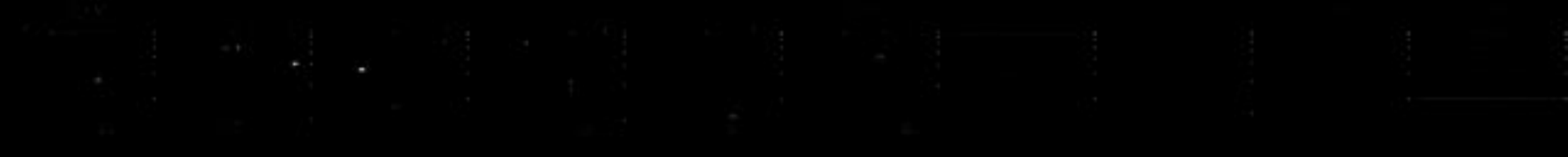}
    \end{subfigure}
    \vskip 0.1in
    \caption{Decoded sampled forward model predictions compared to the true environment state for \textsc{Hero, Qbert, Seaquest} and \textsc{Beam Rider}.}
    \vskip -0.1in
    \label{fig:reconstrpart2}
\end{figure*}

Figures \ref{fig:reconstrpart1} and \ref{fig:reconstrpart2} provide a comparison between a true environment state image $s_{t+1}$ and a decoded state image $\hat{s}_{t+1} = D_\chi(\hat{z}_{t+1})$ after training 1M environment steps, where $\hat{z}_{t+1}$ is the latent state predicted by a sampled forward model $f_{\tilde{\theta}}$ given a state image $s_{t}$ encoded as $z_{t} = E_\chi(s_t)$, action $a_t$, and hidden state $h_t$. Figures \ref{fig:reconstrpart1} and \ref{fig:reconstrpart2} also show the corresponding error image $\hat{s}_{t+1} - s_{t+1}$. As discussed in Section \ref{sec:main-results}, \textsc{Beam Rider} state images contain important details that are missed (see second and third error images in the last row of Fig. \ref{fig:reconstrpart2}).

Additionally, using decoded predictions of sampled forward models we can illustrate the uncertainty quantification over environment models. The following video showcases decoded latent rollouts generated by different sampled forward models, where each rollout was initiated from the same starting state after 75k environment steps of training: \url{https://gifyu.com/image/SIMGF}. The video clearly demonstrates significant diversity in predictions made by distinct sampled models, resulting in disparate trajectories over time, consequently driving the exploration of the agent.

\end{document}